\documentclass[3p,times,authoryear]{elsarticle}

\usepackage[utf8]{inputenc}
\usepackage[T1]{fontenc}
\usepackage{amsmath,amssymb,amsfonts,mathtools}
\usepackage{bm}
\usepackage{graphicx}
\usepackage{subcaption}
\usepackage{booktabs}
\usepackage{multirow}
\usepackage{array}
\usepackage{enumitem}
\usepackage{hyperref}
\usepackage{xcolor}
\usepackage{url}
\usepackage{algorithm}
\usepackage{algpseudocode}
\usepackage{makecell}
\usepackage{pifont}
\journal{Transportation Research Part C: Emerging Technologies}

\newcommand{\ccond}{\mathbf{c}}

\newcommand{\loss}{\mathcal{L}}
\newcommand{\cmark}{\ding{51}}
\newcommand{\xmark}{\ding{55}}

\begin{document}

\begin{frontmatter}

\title{FeaXDrive: Feasibility-aware Trajectory-Centric Diffusion Planning for End-to-End Autonomous Driving}

\author[inst1]{Baoyun Wang\fnref{label1}}
\ead{wangby@tongji.edu.cn}

\author[inst1]{Zhuoren Li\fnref{label1}}
\ead{1911055@tongji.edu.cn}
\fntext[label1]{These authors contributed equally to this work.}

\author[inst1]{Ran Yu}
\ead{2433113@tongji.edu.cn}

\author[inst1]{Yu Che}
\ead{rain_car@tongji.edu.cn}

\author[inst1]{Xinrui Zhang}
\ead{2210805@tongji.edu.cn}

\author[inst1]{Ming Liu}
\ead{2110215@tongji.edu.cn}

\author[inst2]{Jia Hu}
\ead{hujia@tongji.edu.cn}

\author[inst3]{Lv Chen}
\ead{lyuchen@ntu.edu.sg}

\author[inst1]{Bo Leng\corref{cor1}}
\ead{lengbo@tongji.edu.cn}
\cortext[cor1]{Corresponding author}


\affiliation[inst1]{
  organization={College of Automotive and Energy Engineering, Tongji University},
  city={Shanghai},
  postcode={201804},
  country={China}
}
\affiliation[inst2]{
  organization={College of Transportation Engineering, Tongji University},
  city={Shanghai},
  postcode={201804},
  country={China}
}
\affiliation[inst3]{
  organization={School of Mechanical and Aerospace Engineering, Nanyang Technological University},
  postcode={639798},
  country={Singapore}
}

\begin{abstract}
End-to-end diffusion planning has shown strong potential for autonomous driving, but the physical feasibility of generated trajectories remains insufficiently addressed. In particular, generated trajectories may exhibit local geometric irregularities, violate trajectory-level kinematic constraints, or deviate from the drivable area, indicating that the commonly used noise-centric formulation in diffusion planning is not yet well aligned with the trajectory space where feasibility is more naturally characterized. To address this issue, we propose FeaXDrive, a feasibility-aware trajectory-centric diffusion planning method for end-to-end autonomous driving. The core idea is to treat the clean trajectory as the unified object for feasibility-aware modeling throughout the diffusion process. Built on this trajectory-centric formulation, FeaXDrive integrates adaptive curvature-regularized training to improve intrinsic geometric and kinematic feasibility, drivable-area guidance within reverse diffusion sampling to enhance consistency with the drivable area, and feasibility-aware GRPO post-training to further improve planning performance while balancing trajectory-space feasibility. Experiments on the NAVSIM benchmark show that FeaXDrive achieves strong closed-loop planning performance while substantially improving trajectory-space feasibility. These findings highlight the importance of explicitly modeling trajectory-space feasibility in end-to-end diffusion planning and provide a step toward more reliable and physically grounded autonomous driving planners. 
\end{abstract}


\end{frontmatter}

\section{Introduction}

Autonomous driving is expected to play an important role in future intelligent transportation systems \citep{LiTITS2026}. In recent years, End-to-end (E2E) learning paradigm has attracted increasing attention, as it aims to directly map scene observations to driving actions within a unified framework \citep{hu2023uniad}. Representative works such as UniAD, VAD, and related E2E planning methods have demonstrated the potential of unified perception, prediction, and planning for autonomous driving \citep{hu2023uniad,jiang2023vad}. Building on this paradigm, VLM-enhanced E2E approaches further introduce multimodal semantic understanding and reasoning capabilities, thereby improving generalization and enabling better interpretation of complex traffic scenes, long-tail events, and high-level navigation intentions \citep{lmdrive,tian2025drivevlm,autovla}. Representative examples include LMDrive, which explores language-guided closed-loop E2E driving \citep{lmdrive}, and DriveVLM, which demonstrates the potential of VLMs for understanding and planning in complex and long-tail driving scenarios \citep{tian2025drivevlm}. Meanwhile, diffusion-based planning has emerged as an increasingly prominent direction in E2E autonomous driving, owing to its strong capability for modeling multimodal driving behaviors and rich trajectory distributions \citep{zheng2025diffusionplanner,liao2025diffusiondrive}.  

However, the physical feasibility of generated trajectories remains insufficiently addressed in existing E2E planning methods. For example, generated trajectories may exhibit abrupt point-wise discontinuities or unnatural deflections, and may violate kinematic limits when viewed from the perspective of the overall trajectory. Moreover, the generated trajectory, together with the corresponding vehicle spatial occupancy, may deviate from the drivable area, thereby breaking consistency between the trajectory and the geometric constraints of the scene, violating basic road-geometry constraints and potentially compromising traffic-rule compliance and the safety of other road users \citep{dauner2024navsim}. Our reproduced baselines further show that drivable-area non-compliance constitutes a major source of planning failure (Table~\ref{tab:failure_cause_stats}). In essence, these problems are different aspects of the same underlying issue: although the model is capable of generating semantically plausible future trajectories, the resulting trajectories do not satisfy both the intrinsic geometric and kinematic requirements of the trajectory itself and the spatial constraints imposed by the road environment \citep{katrakazas2015motionplanning,song2023motionplanningcontrol,rahman2025trajectory}. We refer to this notion of executability in trajectory space as trajectory-space feasibility.

\begin{table}[t]
\centering
\caption{Failure cause distribution in score-zero planning scenes of reproduced diffusion-based planners. Percentages are computed with respect to the score-zero scenes of each planner. }
\small
\setlength{\tabcolsep}{6pt}
\renewcommand{\arraystretch}{1.08}
\begin{tabular}{lcc}
\toprule
\textbf{Failure Cause} & \textbf{DiffusionDrive} \citep{liao2025diffusiondrive} & \textbf{ReCogDrive} \citep{li2025recogdrive} \\
\midrule
Drivable-area non-compliance & 69.25\% & 56.59\% \\
At-fault collision           & 32.42\% & 45.44\% \\
Both                         & 1.67\%  & 2.03\% \\
\bottomrule
\end{tabular}
\label{tab:failure_cause_stats}
\end{table}

A common formulation in diffusion-based planning is noise-centric parameterization, in which the model predicts the noise term or noise residuals during both training and inference \citep{ho2020ddpm,song2020ddim,karras2022edm}. Although this formulation is effective for general generative tasks, its prediction objective lies in noise space, making it difficult to explicitly represent or effectively characterize the physical feasibility of trajectories. As a result, feasibility-related signals can only influence the generation process indirectly through intermediate variables, making them difficult to incorporate into training and sampling in a stable and precise manner. This separation between the prediction space and the feasibility space results in longer and less direct propagation paths for feasibility-related signals, weaker training supervision, and less intuitive correction during inference, while also reducing the physical interpretability of the overall method.

Based on this observation, we construct E2E diffusion planning in a trajectory-centric manner, such that the future clean trajectory serves as the unified core object in both training and inference. This reformulation is particularly suitable for autonomous driving planning, because unlike high-dimensional generation targets such as natural images, a planned trajectory is low-dimensional, highly structured, and directly tied to physically interpretable planning variables \citep{li2025jit}. Local geometric regularity, curvature-related constraints, kinematic feasibility, and the spatial relationship between the vehicle footprint and the drivable area are all naturally expressed in clean trajectory space rather than noise space \citep{katrakazas2015motionplanning,song2023motionplanningcontrol,rahman2025trajectory}. As a result, directly modeling and optimizing the clean trajectory provides a unified interface for feasibility-aware training and sampling guidance.

\begin{figure}[t]
    \centering
    \includegraphics[width=0.98\linewidth]{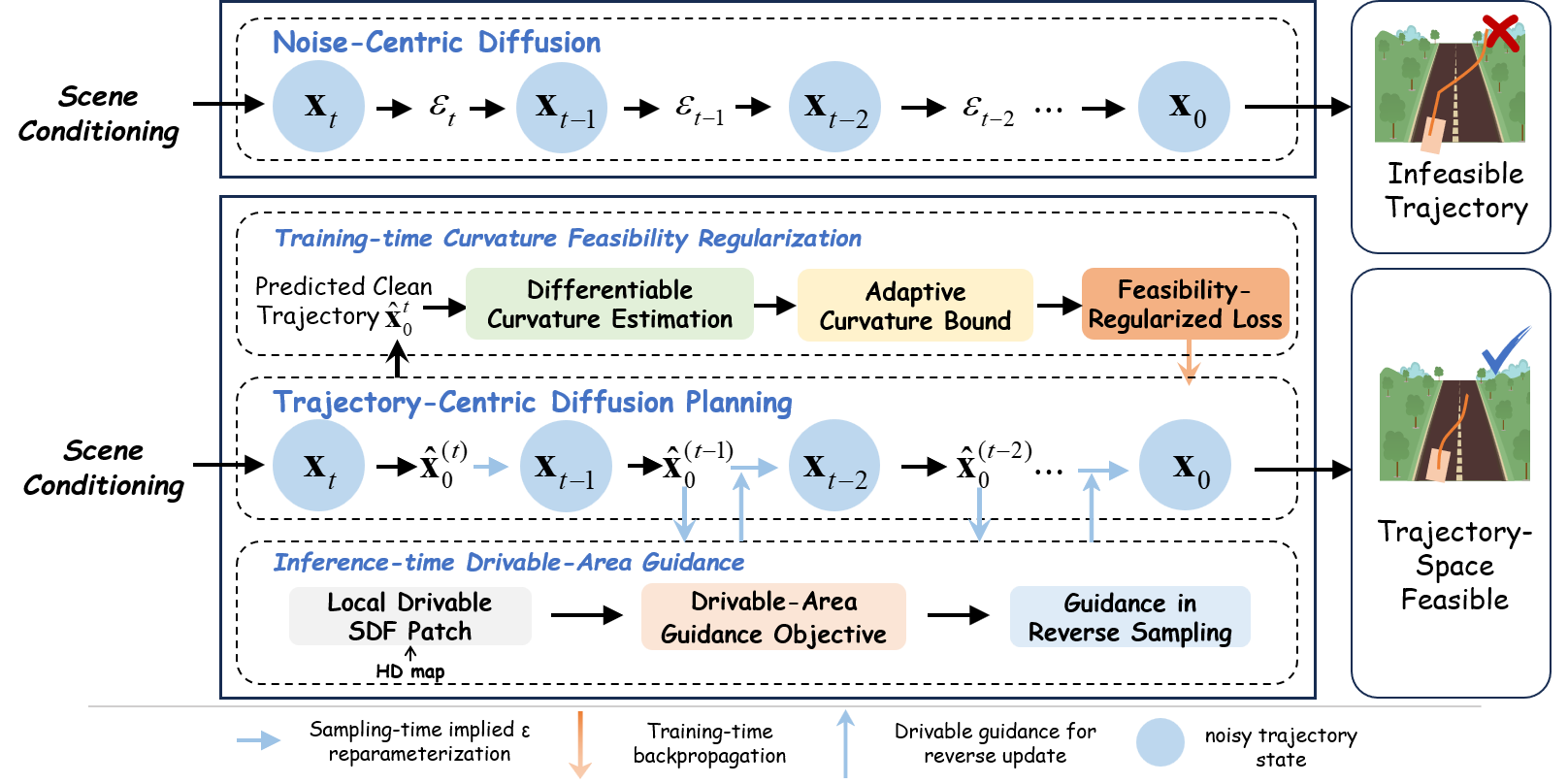}
    \caption{Overview of the proposed FeaXDrive. Compared with noise-centric diffusion planning, FeaXDrive adopts a trajectory-centric formulation in which the predicted clean trajectory serves as the unified object for feasibility-aware modeling. On this basis, the method combines training-time curvature feasibility regularization and inference-time drivable-area guidance to enhance trajectory-space feasibility throughout the diffusion planning process.}
    \label{fig:overview}
\end{figure}

In this work, we propose FeaXDrive. The core idea is to treat the clean trajectory as the explicit carrier of feasibility-related information throughout the diffusion process. By making it the shared optimization object for feasibility-aware modeling throughout training, inference, and post-training, the proposed method provides a unified interface for feasibility enhancement in trajectory space. Specifically, during training, we impose adaptive differentiable curvature constraints directly on the predicted clean trajectory; during inference, we inject drivable-area guidance into the clean trajectory estimated at each reverse sampling step, allowing local road-geometry priors to directly influence trajectory generation; and during post-training, we further incorporate feasibility-aware GRPO fine-tuning to improve planning performance while balancing trajectory-space feasibility. An overview of the proposed method is shown in Fig.~\ref{fig:overview}.

This design allows feasibility-related information to be imposed at the same representation level as the generated trajectory. As a result, FeaXDrive provides a unified basis for training-time feasibility regularization, inference-time guidance, and post-training feasibility-aware optimization, thereby improving trajectory-space feasibility.

Our main contributions are summarized as follows:
\begin{itemize}[leftmargin=1.4em]
    \item \textbf{We propose a trajectory-centric diffusion planning framework for feasibility-aware E2E autonomous driving.}  Different from conventional noise-centric diffusion planners, FeaXDrive treats the predicted clean trajectory as the unified object throughout the diffusion process, which provides a direct trajectory-space interface for feasibility modeling across training, inference, and post-training, where intrinsic trajectory feasibility and drivable-area compliance can be explicitly modeled and optimized.
    \item \textbf{We introduce an adaptive curvature-based feasibility regularization strategy to improve the intrinsic feasibility of generated trajectories.}  Based on differentiable curvature estimation from the predicted clean trajectory, we construct a speed-adaptive curvature bound that combines geometric curvature limits and lateral-acceleration limits. The resulting feasibility-regularized loss is jointly optimized with the trajectory prediction loss, suppressing local trajectory irregularities and curvature spikes during training.
    \item \textbf{We develop a drivable-area guidance mechanism to improve drivable-area compliance during inference.} By constructing a local drivable-area signed distance field and evaluating the vehicle footprint rather than only the trajectory centerline, the proposed guidance objective injects road-geometry priors into each reverse diffusion step. This mechanism progressively steers the diffusion sampling process toward drivable-area-compliant trajectories.
    \item \textbf{We further incorporate feasibility-aware Group Relative Policy Optimization (GRPO) fine-tuning for policy optimization.} By augmenting the evaluation-score-based planning reward with explicit trajectory-space feasibility preferences, the planner is encouraged to generate trajectories that better balance closed-loop planning performance and feasibility, thereby extending feasibility modeling to downstream policy optimization.
    
\end{itemize}

\section{Related Work}

\subsection{{E2E Autonomous Driving Planning}}

Autonomous driving systems were initially developed under modular pipelines, where perception, prediction, planning, and control were treated as separate components \citep{badue2021survey,leng2026seamless,fan2024detection}. As learning-based methods advanced, research gradually moved toward more unified architectures, eventually leading to E2E frameworks that directly map scene observations to planned trajectories. Representative methods such as TransFuser \citep{transfuser}, UniAD \citep{hu2023uniad}, VAD \citep{jiang2023vad}, and Hydra-MDP \citep{hydra_mdp} have demonstrated the potential of integrating perception, prediction, and planning in a single E2E architecture, thereby reducing the complexity of hand-crafted modular pipelines. 

Building on this paradigm, subsequent research introduced large language models and vision-language models into E2E driving systems. Early works such as DriveGPT4 \citep{drivegpt4}, GPT-Driver \citep{gptdriver}, Agent-Driver \citep{agentdriver}, and LMDrive \citep{lmdrive} explored LLMs/VLMs for driving explanation, planning-oriented reasoning, and closed-loop E2E driving, showing that language modeling can provide richer high-level semantic priors than purely vision-based approaches. Subsequent methods, including DriveVLM \citep{tian2025drivevlm}, OmniDrive \citep{omnidrive}, Senna \citep{senna}, and ORION \citep{orion}, further advanced this direction through hierarchical planning, integrated perception-reasoning-planning, language-guided trajectory planning, and vision-language-instructed action generation. More recently, E2E frameworks have moved toward tighter integration of high-level reasoning and action generation, reflecting a broader shift toward more unified generative modeling frameworks for autonomous driving \citep{emma,autovla}. 

\subsection{Diffusion-Based E2E Trajectory Planning}

The standard E2E paradigm based on imitation learning (IL) remains limited in its ability to capture the multimodal distribution of expert behaviors. Diffusion models~\citep{DiffusionD} have therefore emerged as an important generative framework for trajectory generation and autonomous driving planning. Compared with deterministic regression-based methods, diffusion-based planning is better suited to modeling the inherently multimodal nature of autonomous driving and can generate full future trajectories or action sequences through iterative denoising. Foundational diffusion methods~\citep{ho2020ddpm, song2020ddim} provide a general framework for conditional generation and stable sampling, while action-diffusion methods \citep{chi2025diffusionpolicy} further demonstrate the potential of diffusion models for continuous action modeling and policy learning. In autonomous driving and related trajectory-generation tasks, existing diffusion-based methods can be roughly grouped into three lines: multimodal trajectory or behavior distribution modeling, as exemplified by Guided Conditional Diffusion for Controllable Traffic Simulation~\citep{zhong2023ctg} and GoalFlow \citep{xing2025goalflow}; action or policy diffusion, represented by Diffusion Policy \citep{chi2025diffusionpolicy} and DiffE2E \citep{zhao2025diffe2e}, which apply diffusion directly in action space or hybrid action representations; and efficient diffusion planning for autonomous driving, represented by DiffusionDrive \citep{liao2025diffusiondrive}, which improves sampling efficiency through truncated diffusion and anchor priors. In addition, works such as M2Diffuser \citep{yan2025m2diffuser} show that diffusion models can also be combined with explicit trajectory optimization under stronger structural constraints.

Overall, existing diffusion-based trajectory planning has shown strong potential, yet trajectory-space feasibility has received relatively limited attention. Although the broader diffusion literature has explored guidance, constrained sampling, and projection-based refinement \citep{christopher2024pgdm,zheng2025diffusionplanner}, unified feasibility modeling in trajectory space across both training and inference is still lacking in autonomous driving.

\subsection{Feasibility-Enhanced E2E Trajectory Planning}

Existing studies on trajectory feasibility in autonomous driving have explored several directions. Some works improve consistency with scene structure and road geometry through map-aware representations, road-constrained losses, or additional geometric priors, thereby enhancing compliance with the drivable area and lane structure \citep{scenecompliant2020,ellipse2021,laneheading2021}. Other methods adopt a predict-then-constrain paradigm, in which candidate trajectories are first generated by a learning-based model and then refined through external optimizers, MPC, or replacement mechanisms \citep{injectingknowledge2021,safetynet2022,dynamicdrivablecorridor2025}. More recent studies further inject constraints, rewards, or guidance signals directly into generation and policy learning, for example through sampling guidance \citep{diffusiones2024,zheng2025diffusionplanner}, reward modeling \citep{gendrive2025}, aligned policy optimization \citep{diffusiondrivev2,safeauto2025,safevla2025,alphadrive2025}, or human feedback \citep{trajhf2025}.

Despite these advances, existing feasibility-enhanced trajectory planning methods still have several limitations. Many rely on post-generation correction, validation, or replacement rather than unified feasibility modeling of the generation process itself \citep{injectingknowledge2021,safetynet2022,dynamicdrivablecorridor2025}, making feasibility enforcement external and weakly coupled with trajectory generation. In guided generation frameworks, feasibility-related objectives are often introduced only as auxiliary signals rather than being directly integrated into both training and inference \citep{diffusiones2024,zheng2025diffusionplanner,gendrive2025,safevla2025,li2025recogdrive}, which may lead to weaker supervision, less intuitive inference-time correction, and reduced physical interpretability.

\section{Problem Formulation}

\subsection{End-to-End Autonomous Driving Planning}

We formulate E2E autonomous driving as conditional trajectory generation under a diffusion framework. Given a scene condition $\ccond$, which summarizes the driving context including sensor observations, historical ego states, and navigation instructions, the E2E autonomous driving system aims to generate a future motion trajectory. Beyond standard diffusion-based trajectory generation, this work focuses on trajectory-space feasibility, which remains insufficiently addressed in existing diffusion-based E2E approaches.
Formally, we denote the future ego trajectory as:
\begin{equation}
\mathbf{x}_0 =
[\mathbf{s}_1, \mathbf{s}_2, \ldots, \mathbf{s}_H]
\in \mathbb{R}^{H \times d},
\end{equation}
where $\mathbf{s}_i$ denotes the ego waypoint at the $i$-th future time step, $H$ is the planning horizon, and $d$ is the waypoint dimensionality. Each ego waypoint is represented as:
\begin{equation}
\mathbf{s}_i = (p^x_i, p^y_i, p^\theta_i),
\end{equation}
where $(p^x_i, p^y_i)$ denotes the position of the ego pose center, and $p^\theta_i$ denotes the heading angle. Accordingly, the E2E planner aims to learn the conditional distribution $\mathbf{x}_0 \sim p(\mathbf{x}_0 \mid \ccond)$, thereby directly mapping scene understanding to trajectory generation within a unified framework.

\subsection{Diffusion-Based Trajectory Planning}

Under the above E2E autonomous driving formulation, we model the conditional distribution of the future ego trajectory given the scene condition \(\ccond\) using a diffusion generative framework. Let $\mathbf{x}_0$ denote the clean trajectory. In standard diffusion modeling, the forward process progressively injects Gaussian noise into $\mathbf{x}_0$, yielding a noisy intermediate state $\mathbf{x}_t$ at diffusion step $t$:
\begin{equation}
\mathbf{x}_t = \sqrt{\bar{\alpha}_t}\, \mathbf{x}_0 + \sqrt{1-\bar{\alpha}_t}\, \epsilon,
\qquad
\epsilon \sim \mathcal{N}(0, I),
\end{equation}
where $\bar{\alpha}_t$ denotes the cumulative signal-preservation coefficient determined by the noise schedule.
Correspondingly, the reverse diffusion process aims to progressively recover the future ego trajectory from the noisy state, given the current diffusion state $\mathbf{x}_t$, the diffusion step $t$, and the scene condition $\ccond$. A common formulation in diffusion-based trajectory planning is noise-centric parameterization, in which the network is trained to predict the noise term in the current diffusion state:
\begin{equation}
\hat{\epsilon} = f_{\theta}(\mathbf{x}_t, t, \ccond).
\end{equation}

Under this formulation, both the learning objective and the reverse sampling updates are primarily defined in noise space, thereby modeling the conditional trajectory distribution $p(\mathbf{x}_0 \mid \ccond)$. Although diffusion-based planning is effective in modeling multimodal future driving behaviors and generating complete trajectories through iterative denoising, the noise-centric parameterization is not naturally aligned with trajectory-space feasibility, since feasibility-related properties are difficult to explicitly represent or directly optimize in noise space.

To address this issue, in this paper, we further construct a diffusion-based trajectory planning framework in which the clean trajectory is explicitly predicted and serves as the unified object for feasibility-aware modeling throughout training, inference, and post-training. The subsequent Methods section presents the corresponding design in detail and develops training-time feasibility regularization, inference-time geometric guidance, and post-training feasibility-aware optimization within this trajectory-centric framework.

\section{Methods}

\begin{figure}[t]
    \centering
    \includegraphics[width=0.98\columnwidth]{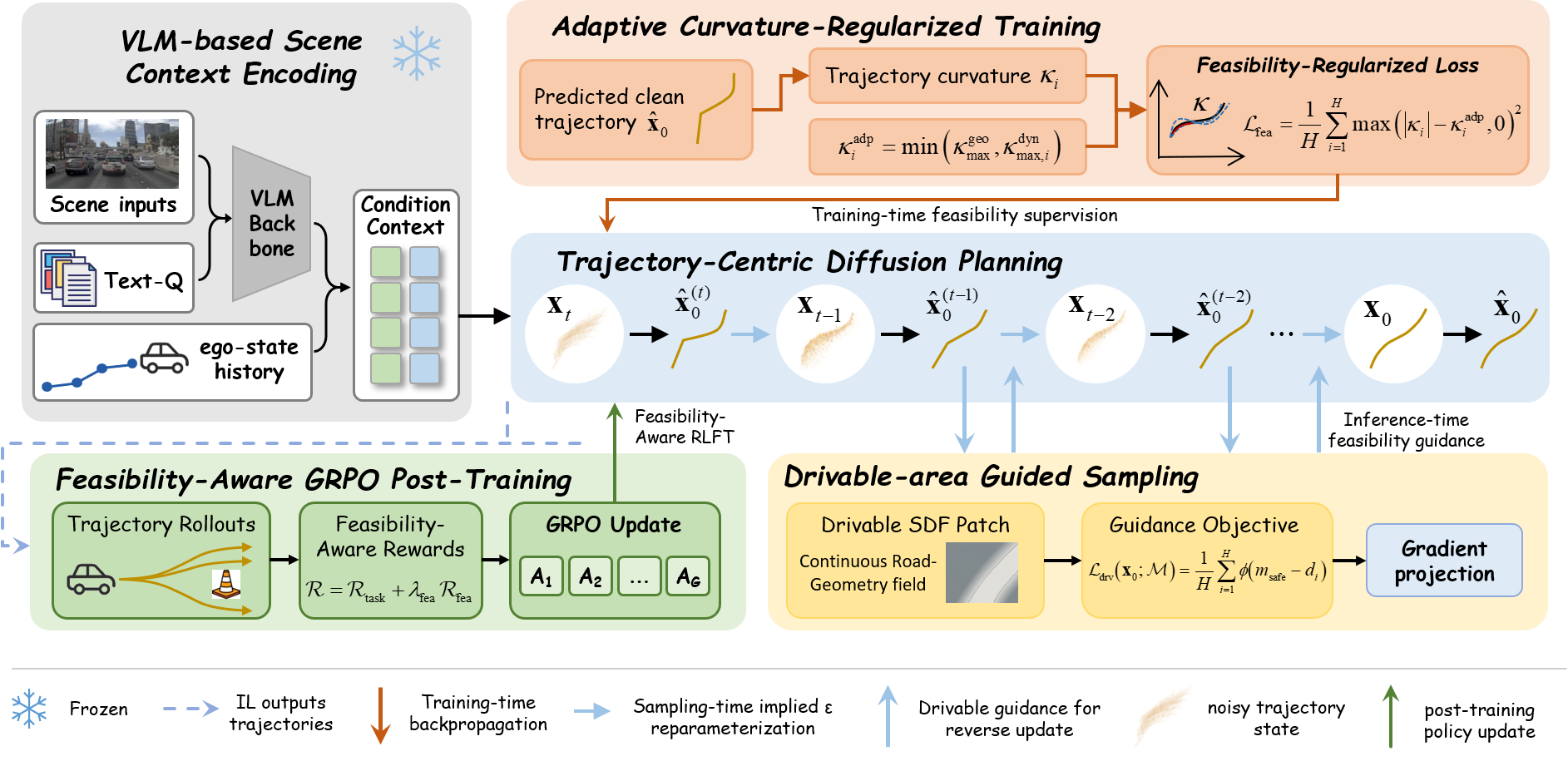}
    \caption{Overall architecture of FeaXDrive. Under a trajectory-centric formulation, the predicted clean trajectory serves as the shared object for feasibility-aware modeling throughout training, inference, and post-training. The method integrates adaptive differentiable curvature-regularized training, drivable-area guidance during reverse diffusion sampling, and feasibility-aware GRPO post-training.}
    \label{fig:architecture}
\end{figure}

This section details the proposed FeaXDrive, a feasibility-aware framework built around a trajectory-centric diffusion planning pipeline. By using the clean trajectory as the explicit interface, FeaXDrive incorporates adaptive differentiable curvature-regularized training for improving intrinsic trajectory feasibility, drivable-area guidance for improving drivable-area consistency during inference, and feasibility-aware GRPO post-training policy optimization for further improving planning performance while balancing trajectory-space feasibility. The overall architecture of FeaXDrive is illustrated in Fig.~\ref{fig:architecture}.

\subsection{Trajectory-Centric Diffusion Planning}

As described in Sec.~3, the planner takes scene images and navigation-related textual instructions as VLM inputs to extract visual-semantic features, which, together with the historical trajectory and ego states, serve as the conditioning information for diffusion planning. Standard diffusion planning typically adopts a noise-centric parameterization, where the network predicts the corresponding noise term $\hat{\epsilon}$ given the noisy trajectory $\mathbf{x}_t$, the diffusion step $t$, and the scene condition $\ccond$. Based on the above discussion, FeaXDrive adopts a trajectory-centric parameterization. Specifically, given the current diffusion state $\mathbf{x}_t$, diffusion step $t$, and scene condition $\ccond$, the planner directly predicts the corresponding clean trajectory estimate:

\begin{equation}
\hat{\mathbf{x}}_0^{(t)} = f_{\theta}(\mathbf{x}_t, t, \ccond),
\end{equation}
where $\hat{\mathbf{x}}_0^{(t)}$ denotes the estimated future trajectory at the current diffusion step, and $f_{\theta}$ is the conditional denoising network. The subsequent feasibility modeling in our method is built on this predicted clean trajectory estimate. During training, predicted clean trajectory estimate $\hat{\mathbf{x}}_0^{(t)}$ is directly supervised by the ground-truth future trajectory $\mathbf{x}^{\text{gt}}_0$; during inference, the current clean trajectory estimate is also explicitly produced at each reverse sampling step. 
Concretely, the basic supervision term in training is defined as
\begin{equation}
\mathcal{L}_{x_0}
=
\left\|
\hat{\mathbf{x}}_0^{(t)}
-
\mathbf{x}_0^{\mathrm{gt}}
\right\|_2^2.
\end{equation}

During inference, at the $t$-th reverse diffusion step, the planner first predicts $\hat{\mathbf{x}}_0^{(t)}$ from the current noisy state $\mathbf{x}_t$, and then uses this clean trajectory estimate to construct the next-state update:
\begin{equation}
\mathbf{x}_{t-1}
=
G\!\left(\mathbf{x}_t, \hat{\mathbf{x}}_0^{(t)}, t\right),
\end{equation}
where $G(\cdot)$ denotes the reverse update operator associated with the specific sampler. 

\subsection{Adaptive Curvature-Regularized Training}

Under the unified trajectory-centric parameterization, we introduce an adaptive curvature-regularized training strategy to improve the intrinsic feasibility of generated trajectories. By imposing explicit regularization in the training stage, the model is encouraged to reduce its tendency to generate trajectories with geometric spikes, local irregularities, or violations of speed-adaptive feasibility bounds.

\subsubsection{Differentiable Curvature Estimation and Adaptive Curvature Bound}

Curvature estimation from discrete waypoints is sensitive to higher-order geometric variations and can be easily affected by local fluctuations and waypoint-level noise. To improve the stability of curvature estimation, we apply a lightweight differentiable smoothing operation to the planar coordinates of the predicted clean trajectory.
Let the discrete planar position sequence of the predicted trajectory be 
\begin{equation}
\hat{\mathbf{P}}_i = (\hat{p}^x_i,\hat{p}^y_i), \quad i = 1,\ldots,H.
\end{equation}
We apply a fixed lightweight 1D convolution kernel to the planar position sequence of the predicted trajectory, yielding a smoothed position trajectory for curvature estimation.
\begin{equation}
\tilde{\mathbf{P}}_i = \mathcal{S}\!\left(\hat{\mathbf{P}}_i\right),
\end{equation}
where $\mathcal{S}(\cdot)$ denotes the differentiable smoothing operator. This operation preserves the global trajectory trend while effectively suppressing local discrete spikes, thereby improving the robustness of subsequent curvature estimation.

After obtaining the smoothed trajectory, we estimate curvature based on arc-length parameterization rather than directly using temporal finite-difference approximation. This is because the spatial spacing between waypoints is not always uniform across different scenarios. Compared with directly differentiating a discrete time sequence, arc-length parameterization is more consistent with the geometric definition of a trajectory and is therefore better suited to characterizing its geometric shape.

First, the arc-length increment between adjacent smoothed waypoints is computed as
\begin{equation}
\Delta \ell_i
=
\max\!\left(
\left\|
\tilde{\mathbf{P}}_{i+1}
-
\tilde{\mathbf{P}}_i
\right\|_2,
\epsilon_\ell
\right),
\quad i=1,\ldots,H-1.
\end{equation}
where $\epsilon_\ell > 0$ is the minimum arc-length constant used to prevent local arc-length degeneration. This gives the cumulative arc-length parameter $\ell_i$. Under the arc-length parameter $\ell$, the trajectory is regarded as a 2D curve $(\tilde{x}(\ell), \tilde{y}(\ell))$, and its first and second derivatives with respect to arc length are estimated. The corresponding curvature is estimated as
\begin{equation}
    \kappa_i =
    \frac{\tilde{x}'(\ell_i)\tilde{y}''(\ell_i)-\tilde{y}'(\ell_i)\tilde{x}''(\ell_i)}
    {\left(\tilde{x}'(\ell_i)^2+\tilde{y}'(\ell_i)^2\right)^{3/2}+\epsilon_{\kappa}},
\end{equation}
where $\epsilon_{\kappa}$ is a numerical stabilizer term. This arc-length-based estimation makes the curvature measure better reflect the intrinsic geometric shape of the trajectory, rather than artifacts introduced by discrete-time parameterization.

A fixed geometric curvature threshold is insufficient to characterize curvature-related feasibility across different motion regimes. In low-speed scenarios, curvature anomalies are mainly associated with local geometric irregularities and discrete waypoint fluctuations. As speed increases, however, even moderate curvature may induce excessive lateral-acceleration demand. Therefore, we construct a speed-aware curvature bound that combines a fixed geometric upper bound with lateral-acceleration limits.

Let the speed at the $i$-th trajectory point be $v_i$, the maximum allowable lateral acceleration be $a_{\max}^{\mathrm{lat}}$, and the fixed geometric curvature upper bound be $\kappa_{\max}^{\mathrm{geo}}$. According to the lateral acceleration relation
\begin{equation}
    \left| a_{\mathrm{lat}} \right| = v^2 \left| \kappa \right|
\end{equation}
the curvature upper bound implied by the lateral dynamic constraint can be derived as
\begin{equation}
    \kappa_{\max,i}^{\mathrm{dyn}} = \frac{a_{\max}^{\mathrm{lat}}}{v_i^2+\epsilon_v},
\end{equation}
where $\epsilon_v$ is a numerical stability term. Furthermore, we define the final curvature bound as
\begin{equation}
    \kappa_i^{\mathrm{adp}} =
    \min\!\left(\kappa_{\max}^{\mathrm{geo}},\ \kappa_{\max,i}^{\mathrm{dyn}}\right)
    =
    \min\!\left(\kappa_{\max}^{\mathrm{geo}},\ \frac{a_{\max}^{\mathrm{lat}}}{v_i^2+\epsilon_v}\right).
\end{equation}

This definition is essentially a hybrid geometric--dynamic constraint, reflecting the automatic switching of the dominant constraint term across different speed regimes. When the vehicle speed is low, the effective threshold is often given by the fixed geometric curvature upper bound $\kappa_{\max}^{\mathrm{geo}}$; as speed increases, the dynamic term gradually becomes more restrictive, causing the effective curvature upper bound to tighten automatically with the motion state, thereby more effectively constraining kinematic curvature violations. As a result, the adopted curvature constraint not only preserves local geometric regularity, but also provides a feasibility constraint that is more consistent with vehicle kinematics.

\subsubsection{Trajectory Feasibility-Aware Loss}

Based on the above curvature estimation and adaptive bound, we define a dynamics-aware training loss to penalize trajectory segments whose curvature exceeds the speed-adaptive curvature bound:
\begin{equation}
    \loss_{\mathrm{cur}}
    =
    \frac{1}{H}\sum_{i=1}^{H}
    \max\!\left(|\kappa_i|-\kappa_i^{\mathrm{adp}},0\right)^2.
\end{equation}
Accordingly, the total training loss can be written as
\begin{equation}
\mathcal{L}_{\mathrm{train}}
=
\mathcal{L}_{x_0}
+
\lambda_{\mathrm{cur}}
\mathcal{L}_{\mathrm{cur}}.
\end{equation}
where $\lambda_{\mathrm{cur}}$ is the weight that balances trajectory supervision and kinematic regularization. The role of $\loss_{\mathrm{cur}}$ is to impose a more physically meaningful bias on the trajectory distribution, thereby suppressing local geometric irregularities and improving trajectory-level kinematic feasibility.

\subsection{Constraint-Aware Inference with Drivable-Area Guidance}

To further improve trajectory-space feasibility, especially the spatial consistency between the generated trajectory and the drivable area, we introduce constraint-aware diffusion sampling with drivable-area guidance during inference. The core idea is that, in the stages of reverse diffusion, the sampler no longer relies only on the model’s generative distribution; instead, local drivable-area geometric priors are injected as guidance into the clean trajectory estimate at each step, enabling online scene-aware geometric correction.

\subsubsection{Guidance within Reverse Diffusion Sampling}

Starting from an initial noisy trajectory \(\mathbf{x}_T \sim \mathcal{N}(0,I)\), 
the reverse diffusion process iteratively performs guidance-aware updates for 
\(t=T,T-1,\dots,1\). At each reverse sampling step \(t\), we first predict a clean trajectory estimate
\(\hat{\mathbf{x}}_0^{(t)}\) from the current diffusion state \(\mathbf{x}_t\). 
Instead of directly using \(\hat{\mathbf{x}}_0^{(t)}\) to generate the next state, 
we apply a constraint-based correction based on the local road-geometry prior 
\(\mathcal{M}\), yielding a guided trajectory estimate:
\begin{equation}
\tilde{\mathbf{x}}_0^{(t)}
=
\mathcal{C}\!\left(\hat{\mathbf{x}}_0^{(t)}; \mathcal{M}\right),
\end{equation}
where \(\mathcal{M}\) denotes the local drivable-area geometric prior, and 
\(\mathcal{C}(\cdot)\) denotes the constraint operator corresponding to the guidance. 
The sampler then continues the reverse update based on the corrected clean trajectory:
\begin{equation}
\mathbf{x}_{t-1}
=
G\!\left(\mathbf{x}_t, \tilde{\mathbf{x}}_0^{(t)}, t\right),
\qquad t=T,\dots,1.
\end{equation}
After the final reverse step, the resulting clean trajectory is taken as the planned trajectory.

The geometric guidance in our method acts on the predicted $\mathbf{x}_0$ at every reverse sampling step and is directly integrated into the reverse sampling chain. It is not an external post-processing step applied after trajectory generation, but an online geometric correction mechanism within the sampling loop. As a result, the guidance directly influences the subsequent evolution of the sampling chain, steering the trajectory toward better consistency with the drivable area during the progressive denoising process.

\subsubsection{Local Drivable-Area SDF Construction}

To introduce drivable-area priors during inference, we define a local drivable-area geometric prior $\mathcal{M}$, which is instantiated in the current implementation as a drivable-area signed distance field (SDF) constructed from a local HD map.

Specifically, we first transform the local road-geometry information into the ego-centric coordinate system with respect to the current ego vehicle, and construct a spatial representation of the drivable region within a finite local window. We then rasterize this local drivable region and compute the corresponding signed distance field. Let the local drivable region be $\mathcal{D} \subset \mathbb{R}^2$, with boundary $\partial \mathcal{D}$. For any point \(\mathbf{q} \in \mathbb{R}^2\) in the local ego-centric plane, we define the signed distance field as
\begin{equation}
S(\mathbf{q}) =
\begin{cases}
\operatorname{dist}(\mathbf{q}, \partial \mathcal{D}), & \mathbf{q} \in \mathcal{D},\\
0, & \mathbf{q} \in \partial \mathcal{D},\\
-\operatorname{dist}(\mathbf{q}, \partial \mathcal{D}), & \mathbf{q} \notin \mathcal{D}.
\end{cases}
\end{equation}
where \(\mathcal{D}\) denotes the drivable region and \(\partial \mathcal{D}\) denotes its boundary. Accordingly, the sign of $S(\mathbf{q})$ characterizes the topological relationship between the point and the drivable region, while its magnitude $\lvert S(\mathbf{q}) \rvert$ explicitly measures the distance from $\mathbf{q}$ to the boundary, i.e., the geometric safety margin or the degree of off-road violation. In particular, when $S(\mathbf{q}) \ge m$, the point lies inside the drivable region and maintains a safety margin of at least $m$ from the boundary.

The SDF transforms the drivable-area constraint from a discrete inside/outside binary test into a continuous geometric distance signal. This not only enables us to determine whether a trajectory goes off-road, but also quantifies its safety margin and degree of boundary violation, thereby providing stable and differentiable geometric information for subsequent gradient-based guidance.

It should be noted that, in our problem formulation, $\mathcal{M}$ is regarded as a unified interface for local road-geometry priors, rather than being tied to any specific map representation. In our current experiments, $\mathcal{M}$ is instantiated from a local HD map, but it can also be replaced in the future by local geometric information provided by lightweight maps, online mapping, or implicit road priors.

\subsubsection{Footprint-Level Drivable-Area Guidance}

Applying drivable-area constraints only to the trajectory center point can easily overlook the relationship between the actual occupied region of the vehicle and the road boundary. In particular, in turning, near-boundary driving, or narrow-road scenarios, the fact that the center point remains inside the road does not necessarily imply that the entire vehicle stays within the drivable region. Therefore, we adopt a footprint-level geometric constraint.

For the vehicle waypoint state at future time step $i$, we construct a rectangular vehicle footprint according to the vehicle length and width. Let the relative coordinates of the four footprint corners in the vehicle coordinate system be $\{\delta_j\}_{j=1}^{4}$. Their positions in the local ego-centric plane are then given by
\begin{equation}
\mathbf{p}_{i,j} =
\begin{bmatrix}
 p^x_i \\
 p^y_i
\end{bmatrix}
+ R(p^\theta_i)\,\delta_j,
\qquad j = 1, \ldots, 4,
\end{equation}
where $R(p^\theta_i)$ denotes the 2D rotation matrix determined by the heading angle $p^\theta_i$. We then perform bilinear sampling on the local SDF at these continuous coordinates to obtain the corresponding signed distance at each corner point:
\begin{equation}
d_{i,j} = S(\mathbf{p}_{i,j}),
\qquad j = 1, \ldots, 4.
\end{equation}

This footprint-based sampling strategy better reflects the true occupied geometry of the vehicle than a center-point constraint. On the one hand, it can explicitly capture cases where a vehicle corner leaves the drivable region while the center point still remains inside. On the other hand, since SDF sampling is continuous and differentiable, gradients can be propagated directly to the vehicle position and heading, enabling more refined geometric guidance.

After obtaining the footprint-level distances, we define a soft barrier guidance objective with a safety-margin term to measure the consistency between the current clean trajectory and the drivable region:
\begin{equation}
\mathcal{L}_{\mathrm{drv}}
\!\left(\hat{\mathbf{x}}_0^{(t)}; \mathcal{M}\right)
=
\frac{1}{4H}
\sum_{i=1}^{H}
\sum_{j=1}^{4}
\phi\!\left(m_{\mathrm{safe}} - d_{i,j}\right),
\end{equation}
where $m_{\mathrm{safe}}$ denotes the desired safety margin to the boundary, and $\phi(\cdot)$ is implemented as a softplus barrier function. When the minimum footprint distance is sufficiently large, the guidance objective approaches zero; when the footprint approaches the boundary or leaves the drivable region, it increases rapidly.

To reduce unnecessary interference with reasonable trajectories, we adopt a trigger mechanism based on whether the footprint goes out of bounds. Guidance is activated only when the footprint points of the current predicted trajectory leave the drivable region or come excessively close to the boundary. For triggered samples, at sampling step \(t\), we perform one or multiple updates 
along the gradient direction of this objective with respect to 
\(\hat{\mathbf{x}}_0^{(t)}\), yielding the guided clean trajectory estimate:
\begin{equation}
\tilde{\mathbf{x}}_0^{(t)}
=
\hat{\mathbf{x}}_0^{(t)}
-
\eta_t
\Phi\!\left(
\nabla_{\hat{\mathbf{x}}_0^{(t)}}
\mathcal{L}_{\mathrm{drv}}
\right),
\end{equation}
where $\eta_t$ denotes the guidance step size at sampling step $t$, and $\Phi(\cdot)$ denotes the gradient normalization and scale modulation operator. In practice, we normalize the gradients on the positional dimensions to avoid excessively large variations in SDF gradient scales across different scenes. In addition, an independent scaling factor can be introduced for the heading dimension to balance geometric correction and trajectory smoothness.

In summary, drivable-area guidance performs online progressive correction on the clean trajectory estimate at each sampling step using the local drivable-area geometry of the current scene, gradually steering it away from non-drivable regions and back toward drivable-area-compliant trajectory generation.

Compared with the adaptive curvature-regularized training, drivable-area guidance provides scene-specific geometric correction for the current scene. The former mainly improves the intrinsic feasibility of trajectories, while the latter mainly enhances spatial consistency with the drivable region. By combining these two modules under a unified trajectory-centric representation, we develop a feasibility-aware diffusion planning method for autonomous driving trajectory planning.

\subsection{Reinforcement Learning Fine-Tuning with Feasibility-Aware GRPO}

Starting from the supervised diffusion planner, we further fine-tune the diffusion planning policy using reinforcement learning (RL) \citep{jinRL} to improve closed-loop planning performance beyond imitation learning while preserving trajectory-space feasibility. Specifically, we instantiate this RL fine-tuning stage with the proposed Feasibility-Aware GRPO. Following the group-relative policy optimization principle \citep{shao2024deepseekmath,guo2025deepseekr1}, Feasibility-Aware GRPO adapts relative policy optimization to the diffusion generation chain and explicitly incorporates trajectory-space feasibility into the reward design. The core idea is to augment the evaluation-score-based planning reward with feasibility preferences, so that policy optimization favors candidate trajectories that achieve both strong benchmark performance and high trajectory-space feasibility.

Specifically, given a scene condition \(\ccond\), the current policy samples a group of candidate trajectories
\begin{equation}
\left\{ \mathbf{x}_{0}^{(g)} \right\}_{g=1}^{G},
\qquad
\mathbf{x}_{0}^{(g)}
\sim
\pi_{\theta}(\cdot \mid \ccond),
\end{equation}
and obtains the corresponding rewards through the planning evaluator 
\begin{equation}
r_g
=
\mathcal{R}\!\left(\mathbf{x}_{0}^{(g)}, \ccond\right).
\end{equation}
Here, \(\mathcal{R}(\cdot)\) jointly characterizes task quality and feasibility preference, and is defined as
\begin{equation}
\mathcal{R}\left(\mathbf{x}_{0}, \ccond\right)
=
\mathcal{R}_{\mathrm{task}}\left(\mathbf{x}_{0}, \ccond\right)
+
\lambda_{\mathrm{fea}}
\mathcal{R}_{\mathrm{fea}}\left(\mathbf{x}_{0}, \ccond\right),
\end{equation}
where $\mathcal{R}_{\mathrm{task}}$ measures the task-level planning quality, $\mathcal{R}_{\mathrm{fea}}$ represents the trajectory feasibility preference, and $\lambda_{\mathrm{fea}}$ is the trade-off coefficient between the two terms. As a result, the reward optimized by GRPO is no longer determined solely by benchmark-oriented performance, but instead explicitly incorporates a preference modeling for trajectory-space feasibility. In this work, $\mathcal{R}_{\mathrm{fea}}$ is characterized by the speed-adaptive curvature feasibility defined in Section~4.2. Specifically, we directly adopt the curvature estimation method and the speed-adaptive constraint criterion introduced in Section~4.2, and incorporate them into the post-training reward design. If a generated trajectory satisfies the curvature-feasibility requirement defined in Section~4.2, it receives a higher feasibility reward; otherwise, its feasibility reward is reduced accordingly, causing it to be disadvantaged in the within-group comparison. In this way, the trajectory-space feasibility modeling introduced above further enters the post-training policy optimization process through feasibility-aware reward shaping.

After obtaining the rewards of \(G\) candidate trajectories for the same scene, GRPO performs relative normalization of the rewards within the group to construct the relative advantage of each trajectory:
\begin{equation}
A_g = \frac{r_g - \mu_r}{\sigma_r + \epsilon_r},
\end{equation}
where \(\mu_r\) and \(\sigma_r\) denote the mean and standard deviation of the rewards within the same group, respectively, and \(\epsilon_r\) is a numerical stability term. Since the diffusion planner generates the final trajectory through a multi-step reverse denoising process, policy optimization does not act only on the terminal trajectory variable, but on the entire denoising chain that generates the trajectory. Accordingly, the GRPO policy optimization term can be written as
\begin{equation}
\mathcal{L}_{\mathrm{RL}}
=
-\mathbb{E}\!\left[
A_g
\sum_{t=1}^{T}
w_t
\log
\pi_{\theta}\!\left(
\mathbf{x}_{t-1}^{(g)}
\mid
\mathbf{x}_{t}^{(g)}, \ccond
\right)
\right],
\end{equation}
where \(w_t\) denotes the denoising-step-related weight. This objective uses the trajectory-level relative advantage to weight the log-likelihood terms along the reverse denoising chain, encouraging the policy to increase the likelihood of candidate trajectories that achieve both high task rewards and favorable feasibility scores. In our implementation, we adopt a single-update GRPO objective for diffusion denoising chains. Each sampled group is used for one policy update, and we do not perform multi-epoch PPO-style reuse of the same samples. Therefore, we do not introduce a clipped likelihood-ratio objective with respect to a separate old policy. The resulting optimization reduces to an advantage-weighted log-likelihood objective over the reverse denoising chain.

In addition, to prevent RL post-training from deviating excessively from the policy prior learned during the IL stage, we introduce a behavior-cloning regularization term based on a fixed reference policy \(\pi_{\mathrm{ref}}\):
\begin{equation}
\mathcal{L}_{\mathrm{BC}}
=
-
\mathbb{E}_{\ccond,\,\mathbf{x}_{1:T}^{\mathrm{ref}} \sim \pi_{\mathrm{ref}}}
\left[
\sum_{t=1}^{T}
\log
\pi_{\theta}
\!\left(
\mathbf{x}_{t-1}^{\mathrm{ref}}
\mid
\mathbf{x}_{t}^{\mathrm{ref}}, \ccond
\right)
\right].
\end{equation}
Here, \(\pi_{\mathrm{ref}}\) is the fixed IL policy, and
\(\mathbf{x}_{T}^{\mathrm{ref}} \rightarrow \cdots \rightarrow \mathbf{x}_{0}^{\mathrm{ref}}\)
denotes the reference denoising chain. The final post-training objective is
\begin{equation}
\mathcal{L}
=
\mathcal{L}_{\mathrm{RL}}
+
\lambda_{\mathrm{BC}} \mathcal{L}_{\mathrm{BC}}.
\end{equation}

Overall, as the post-training stage of this work, Feasibility-Aware GRPO strengthens the model’s preference for high-quality and feasible trajectories from the perspective of policy optimization through feasibility-aware reward shaping. Therefore, the proposed trajectory-centric diffusion planner not only incorporates trajectory-space feasibility modeling during the IL training stage and the sampling stage, but also remains compatible with subsequent policy optimization, thereby further improving the overall balance between benchmark performance and trajectory-space feasibility.

\section{Experiments}

This section evaluates the proposed method in terms of standard planning performance, trajectory-space feasibility, module-wise contributions, and inference efficiency. We first present the experimental setup, then report the main results, followed by ablation and feasibility analysis, efficiency evaluation, and qualitative visualization.

\subsection{Experimental Setup}

\subsubsection{Dataset and Benchmark}
We conduct experiments on the NAVSIM benchmark \citep{dauner2024navsim}, a planning-oriented autonomous driving dataset built on OpenScene \citep{openscene2024}, which is a redistribution of nuPlan \citep{karnchanachari2024nuplan}. NAVSIM provides multimodal driving observations together with closed-loop evaluation for end-to-end autonomous driving planning, and is split into navtrain (1,192 training scenes) and navtest (136 evaluation scenes). In each scene, the planner takes scene images, historical ego states, and navigation-related context as input, and predicts the future ego trajectory. Following the standard NAVSIM evaluation protocol, we report both benchmark planning metrics and the trajectory-space feasibility metrics considered in this work.

\subsubsection{Implementation Details}

Our model follows a VLM-conditioned diffusion planning pipeline and consists of two main components: a vision-language encoder and a diffusion-based trajectory planner. Given a front-view scene image, navigation-related textual instructions, and historical ego states, the vision-language encoder first extracts semantic driving-context representations from the visual and textual inputs. These VLM hidden states, together with historical ego-state features and diffusion-timestep embeddings, form the condition context \(\ccond\) for the diffusion planner. The diffusion planner is implemented as a DiT-based trajectory denoising network \citep{peebles2023scalable}, which takes the noisy trajectory state and \(\ccond\) as input and directly predicts the clean future ego trajectory under a trajectory-centric diffusion parameterization. The trajectory denoising network adopts a DiT architecture, with \(8\) attention heads and \(16\) transformer layers. Based on this pipeline, FeaXDrive introduces adaptive curvature-regularized training during IL training, drivable-area guidance during reverse diffusion sampling, and feasibility-aware GRPO during post-training.

To define the curvature bound used in both training and evaluation, we align the geometric curvature limit with the minimum turning radius of the Chrysler Pacifica, the data-collection vehicle used in NAVSIM. According to the official specification, the minimum turning radius of the Chrysler Pacifica is approximately \(19.8\,\mathrm{ft}\) (\(\approx 6.0\,\mathrm{m}\)) \citep{chrysler2026pacifica_specs}, which, under a low-speed geometric approximation, corresponds to $\kappa_{\mathrm{geo}} \approx 1/{R_{\min}} = {1}/{6.0} \approx 0.166 \,\mathrm{m}^{-1}.$
Accordingly, we set the fixed geometric curvature bound to \(\kappa_{\mathrm{geo}}=0.166\,\mathrm{m}^{-1}\), and set the maximum allowable lateral acceleration to \(a_{\max}^{\mathrm{lat}}=6\,\mathrm{m/s}^2\) (\(\approx 0.61g\)) as an upper bound for near-limit maneuvering. Together, these values are used to construct the speed-aware curvature bound that combines a fixed geometric upper bound with lateral-acceleration limits.

The IL-based diffusion model is first trained for \(100\) epochs on \(4\) A800 GPUs, using bf16 mixed precision and distributed data parallel (DDP), with a per-GPU batch size of \(32\) and a total batch size of \(128\). The vision-language encoder is initialized from the same pretrained VLM checkpoint as ReCogDrive \citep{li2025recogdrive}, based on InternVL3-2B \citep{chen2024internvl}, and its parameters are frozen during training. 

We further conduct a feasibility-aware GRPO fine-tuning stage. Starting from the IL-trained model, we perform GRPO fine-tuning for \(1\) epoch on \(8\) A800 GPUs with a per-GPU batch size of \(8\). Unlike score-only post-training, the GRPO reward incorporates curvature-violation terms together with benchmark score terms, thereby further improving planning performance while balancing trajectory-space feasibility.

During inference, we use DDIM sampling for all variants. Unless otherwise specified, all internal variants share the same inference settings.

\subsubsection{Evaluation Metrics}

We evaluate our method using two groups of metrics:
\begin{itemize}[leftmargin=1.4em]
    \item \textbf{Standard planning metrics:} For the NAVSIM benchmark, we follow the official Predictive Driver Model Score (PDMS), including no-at-fault collision (NC), drivable-area compliance (DAC), time-to-collision (TTC), comfort (Comf.), and ego progress (EP). Higher PDMS indicates better overall planning performance.

    \item \textbf{Trajectory-space feasibility metrics:} To evaluate the main focus of this work, we further consider trajectory-space feasibility. Specifically, we report the curvature violation rate under the adaptive curvature bound and the drivable-area violation rate. Let \(N\) denote the total number of evaluated planning scenes. For scene \(i\), let \(\mathbb{I}^{(i)}_{\kappa}=1\) if the generated trajectory violates the adaptive curvature bound, and \(\mathbb{I}^{(i)}_{\mathrm{drv}}=1\) if the generated trajectory violates the drivable-area constraint. The two rates are defined as
    \begin{equation}
    \mathrm{Curvature\ Viol.\ Rate}
    =
    \frac{1}{N}\sum_{i=1}^{N}\mathbb{I}^{(i)}_{\kappa},
    \qquad
    \mathrm{Drivable\ Area\ Viol.\ Rate}
    =
    \frac{1}{N}\sum_{i=1}^{N}\mathbb{I}^{(i)}_{\mathrm{drv}}.
    \end{equation}
    The former reflects the intrinsic geometric and kinematic aspect of trajectory-space feasibility, while the latter reflects consistency with the drivable area. Together, these metrics evaluate whether the generated trajectories remain feasible in trajectory space while maintaining competitive benchmark performance.

\end{itemize}

\subsection{Main Results}

Table~\ref{tab:navsim_main_results} reports the main closed-loop planning results on the NAVSIM benchmark. Overall, the proposed method achieves strong planning performance under both imitation learning (IL) and reinforcement learning fine-tuning (RLFT). Under the IL setting, FeaXDrive-IL achieves a PDMS of 88.7, outperforming all other IL baselines, including DiffusionDrive (\(88.1\)), WoTE (\(88.3\)), and ReCogDrive-IL (\(86.8\)). In particular, our method achieves the highest DAC (\(97.5\)) and the highest ego progress (\(83.3\)) among the IL methods, indicating that the proposed method remains highly competitive on the standard NAVSIM benchmark even before RLFT.

After RLFT, FeaXDrive further improves the PDMS to 90.0, demonstrating that the proposed planner remains compatible with downstream policy optimization. Although its final PDMS is slightly lower than that of ReCogDrive w/GRPO (\(90.5\)), our method achieves a substantially higher DAC (\(98.3\) vs.\ \(96.7\)), indicating better compliance with the drivable area. Overall, these results show that the proposed method maintains competitive benchmark performance while exhibiting stronger drivable-area consistency.

\begin{table*}[t]
\centering
\caption{Main closed-loop planning results on the NAVSIM benchmark. RLFT denotes reinforcement learning fine-tuning; for FeaXDrive, it is instantiated as feasibility-aware GRPO.}
\small
\setlength{\tabcolsep}{5pt}
\renewcommand{\arraystretch}{1.05}
\begin{tabular}{c|l|c c c c|c|c}
\toprule
\textbf{Method type} & \textbf{Model} & \textbf{NC} $\uparrow$ & \textbf{EP} $\uparrow$ & \textbf{Comf.} $\uparrow$ & \textbf{TTC} $\uparrow$ & \textbf{DAC} $\uparrow$ & \textbf{PDMS} $\uparrow$ \\
\midrule
\multirow{11}{*}{IL}
& VADv2 \citep{vadv2}                   & 97.2 & 76.0 & \textbf{100}  & 91.6 & 89.1 & 80.9 \\
& Driving-GPT \citep{drivinggpt}        & \textbf{98.9} & 79.7 & 95.6 & \textbf{94.9} & 90.7 & 82.4 \\
& Hydra-MDP \citep{hydra_mdp}           & 97.9 & 77.6 & \textbf{100}  & 92.9 & 91.7 & 83.0 \\
& UniAD \citep{hu2023uniad}                   & 97.8 & 78.8 & \textbf{100}  & 92.9 & 91.9 & 83.4 \\
& PARA-Drive \citep{paradrive}          & 97.9 & 79.3 & 99.8 & 93.0 & 92.4 & 84.0 \\
& TransFuser \citep{transfuser}         & 97.7 & 79.2 & \textbf{100}  & 92.8 & 92.8 & 84.0 \\
& DRAMA \citep{drama}                   & 98.0 & 80.1 & \textbf{100}  & 94.8 & 93.1 & 85.5 \\
& ReCogDrive-IL \citep{li2025recogdrive}      & 98.3 & 81.1 & \textbf{100}  & 94.3 & 95.1 & 86.8 \\
& DiffusionDrive \citep{liao2025diffusiondrive} & 98.2 & 82.2 & \textbf{100}  & 94.7 & 96.2 & 88.1 \\
& WoTE \citep{wote}                     & 98.5 & 81.9 & 99.9 & \textbf{94.9} & 97.3 & 88.3 \\
& \textbf{FeaXDrive-IL (Ours)}                   & 98.1 & \textbf{83.3} & \textbf{100}  & 93.6 & \textbf{97.5} & \textbf{88.7} \\
\midrule
\multirow{3}{*}{IL+RLFT}
& TransFuser w/GRPO \citep{transfuser}  & 98.0 & \textbf{88.5} & \textbf{100} & \textbf{96.6} & 94.7 & 87.9 \\
& ReCogDrive w/GRPO \citep{li2025recogdrive}  & 98.1 & 85.9 & \textbf{100} & 95.0 & 96.7 & \textbf{90.5} \\
& \textbf{FeaXDrive (Ours)}               & \textbf{98.2} & 84.2 & \textbf{100} & 94.7 & \textbf{98.3} & 90.0 \\
\bottomrule
\end{tabular}
\label{tab:navsim_main_results}
\end{table*}

To further evaluate the main focus of this work, Table~\ref{tab:feasibility_results} reports the curvature violation rate under a unified evaluation pipeline. We compare our method with two reproduced diffusion-based baselines, namely DiffusionDrive \citep{liao2025diffusiondrive} and ReCogDrive \citep{li2025recogdrive}. Since such feasibility metrics are generally not reported in prior work and are not always directly available from released checkpoints or evaluation code, we reproduce these representative baselines to enable a fair comparison under the same evaluation setup. Under IL training, FeaXDrive-IL achieves a curvature violation rate of only \(0.88\%\), substantially lower than DiffusionDrive (\(8.59\%\)) and ReCogDrive-IL (\(8.05\%\)). After Feasibility-Aware GRPO fine-tuning, FeaXDrive maintains a low curvature violation rate of \(2.40\%\), while ReCogDrive w/GRPO increases to \(15.5\%\). These results suggest that score-oriented post-training may improve benchmark metrics at the cost of trajectory-space feasibility, whereas the proposed feasibility-aware design is able to preserve much stronger curvature feasibility while still achieving competitive benchmark performance.

\begin{table}[t]
\centering
\caption{Comparison of curvature violation rates among reproduced diffusion-based planners.}
\footnotesize
\setlength{\tabcolsep}{4pt}
\renewcommand{\arraystretch}{1.12}
\begin{tabular}{l|c|c|c|c|c}
\toprule
\textbf{Method}
& \makecell[c]{\textbf{DiffusionDrive}\\ \citep{liao2025diffusiondrive}}
& \makecell[c]{\textbf{ReCogDrive-IL}\\ \citep{li2025recogdrive}}
& \makecell[c]{\textbf{ReCogDrive w/GRPO}\\ \citep{li2025recogdrive}}
& \makecell[c]{\textbf{FeaXDrive-IL}\\ \textbf{(Ours)}}
& \makecell[c]{\textbf{FeaXDrive w/FA-GRPO}\\ \textbf{(Ours)}} \\
\midrule
\makecell[l]{\textbf{Curvature Viol. Rate}$\downarrow$}
& 8.59\%
& 8.05\%
& 15.5\%
& \textbf{0.88\%}
& \textbf{2.40\%} \\
\bottomrule
\end{tabular}
\label{tab:feasibility_results}
\end{table}

Overall, the results demonstrate that FeaXDrive not only delivers strong planning performance on NAVSIM, but also enhances trajectory-space feasibility. Across both IL training and post-training fine-tuning, the proposed method consistently improves planning performance while balancing trajectory-space feasibility, validating the effectiveness of the proposed design.

\subsection{Ablation and Feasibility Analysis}

\begin{figure}[t]
    \centering
    \includegraphics[width=0.62\columnwidth]{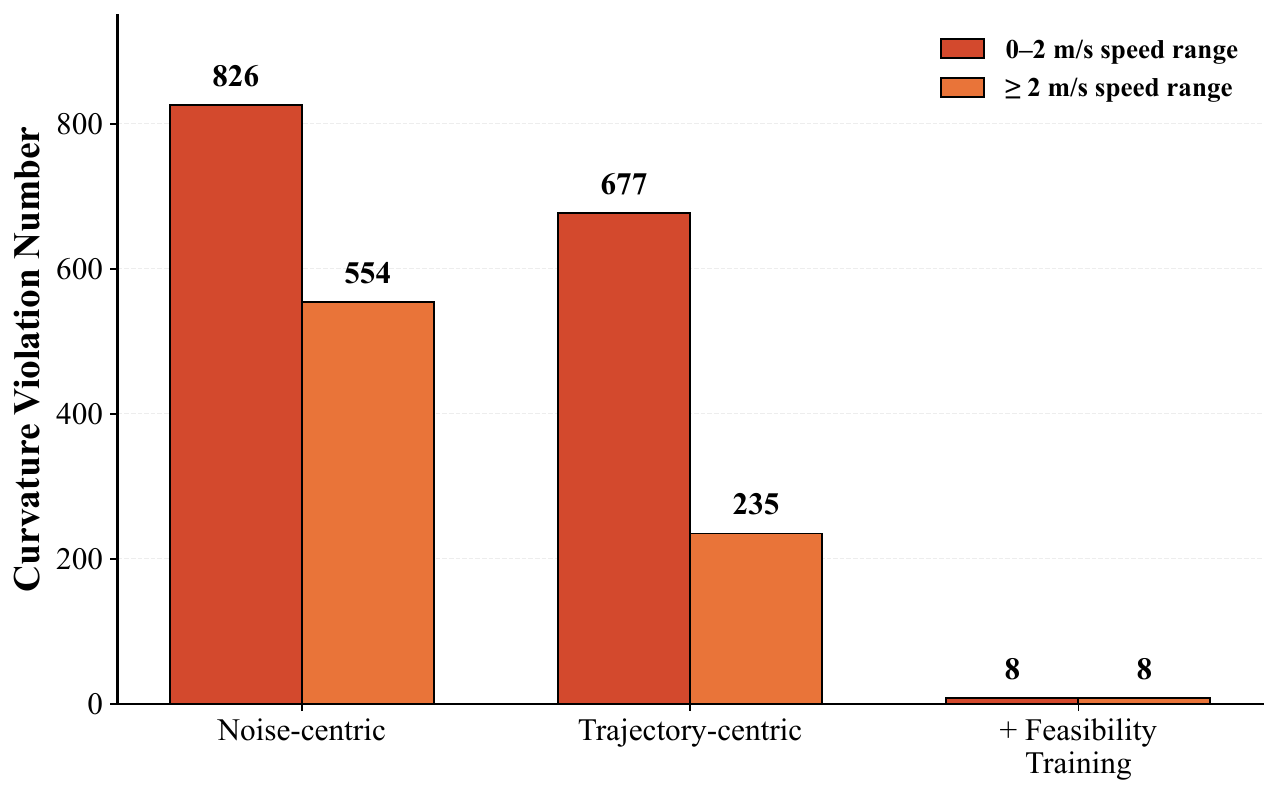}
    \caption{Comparison of curvature violation counts under different IL-stage ablation settings.}
    \label{fig:curvature_violation}
\end{figure}

\begin{figure}[t]
    \centering
    \includegraphics[width=0.45\columnwidth]{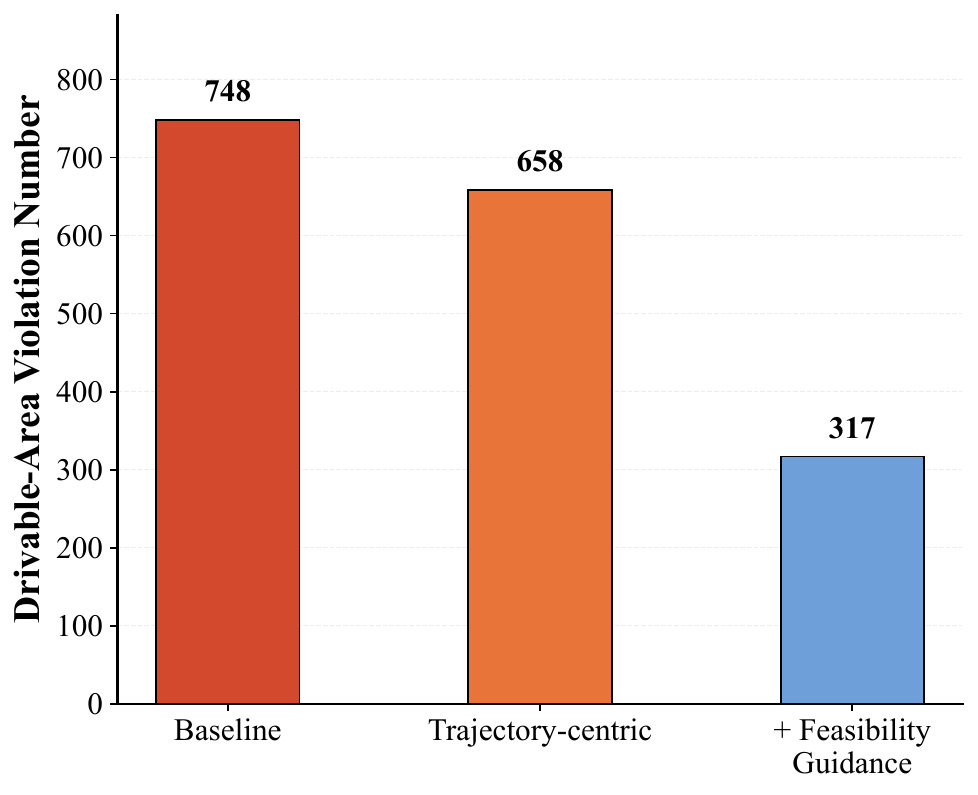}
    \caption{Comparison of drivable-area violation counts under different IL-stage ablation settings.}
    \label{fig:drivable_violation}
\end{figure}

We further analyze the contribution of each module in the proposed method from the perspectives of benchmark performance and trajectory-space feasibility. Table~\ref{tab:ablation_il} reports the IL-stage ablation results, while Table~\ref{tab:ablation_rlft} compares different post-training strategies. Figs.~\ref{fig:curvature_violation} and \ref{fig:drivable_violation} further visualize the effect of adaptive curvature-regularized training and drivable-area guidance on representative feasibility indicators.

\begin{table*}[t]
\centering
\caption{IL-stage ablation study of the proposed method on benchmark performance and trajectory-space feasibility.}
\footnotesize
\setlength{\tabcolsep}{5pt}
\renewcommand{\arraystretch}{1.12}
\begin{tabular}{l|ccc|ccc|cc}
\toprule
\textbf{Method} 
& \makecell[c]{\textbf{x0-pred}}
& \makecell[c]{\textbf{Curvature}\\ \textbf{Regularization}}
& \makecell[c]{\textbf{Sampling}\\ \textbf{Guidance}}
& \textbf{PDMS} $\uparrow$
& \textbf{EP} $\uparrow$
& \textbf{DAC} $\uparrow$
& \makecell[c]{\textbf{Drivable Area}\\ \textbf{Viol. Rate} $\downarrow$}
& \makecell[c]{\textbf{Curvature}\\ \textbf{Viol. Rate} $\downarrow$} \\
\midrule
Baseline               & \xmark & \xmark & \xmark & 85.32 & 79.56 & 93.84 & 6.16\% & 11.36\% \\
Trajectory-centric     & \cmark & \xmark & \xmark & 86.56 & 81.26 & 94.58 & 5.42\% & 7.51\% \\
\makecell[l]{+ Training-time feasibility\\ enhancement} 
& \cmark & \cmark & \xmark & 86.57 & 81.32 & 94.94 & 5.06\% & \textbf{0.13\%} \\
FeaXDrive-IL           & \cmark & \cmark & \cmark & \textbf{88.75} & \textbf{83.34} & \textbf{97.46} & \textbf{2.54\%} & 0.88\% \\
\bottomrule
\end{tabular}
\label{tab:ablation_il}
\end{table*}

\textbf{Effect of the trajectory-centric diffusion planning.}
Comparing the noise-centric diffusion planning baseline with the trajectory-centric diffusion planning method in Table~\ref{tab:ablation_il}, replacing the noise-centric formulation with the trajectory-centric one consistently improves both planning performance and trajectory-space feasibility. Specifically, PDMS increases from \(85.32\) to \(86.56\), ego progress rises from \(79.56\) to \(81.26\), and DAC improves from \(93.84\) to \(94.58\). Meanwhile, the curvature violation rate decreases from \(11.36\%\) to \(7.51\%\), and the drivable-area violation rate drops from \(6.16\%\) to \(5.42\%\). These results indicate that trajectory-centric diffusion planning not only improves trajectory planning performance, but also provides a suitable and unified modeling foundation for subsequent feasibility-aware training and inference in trajectory space.

\textbf{Effect of adaptive curvature-regularized training.}
Adding adaptive curvature-regularized training on top of the trajectory-centric formulation produces the most significant gain in curvature-related feasibility. As shown in Table~\ref{tab:ablation_il}, the curvature violation rate drops from \(7.51\%\) to \(0.13\%\), while PDMS, EP, DAC, and drivable-area violation rate change only marginally. This indicates that the proposed training strategy mainly improves the intrinsic geometric and kinematic feasibility of generated trajectories, without sacrificing benchmark performance. 

This trend is further confirmed by Fig.~\ref{fig:curvature_violation}, where the number of curvature-violation instances is substantially reduced in both speed ranges. We report the statistics in the 0-2 m/s and $\geq 2$ m/s ranges because curvature violations may exhibit different characteristics across motion regimes. In lower-speed cases, curvature anomalies are more likely to be influenced by local geometric irregularities and discretization effects, whereas with increasing speed, curvature violations tend to become more closely associated with trajectory-level kinematic feasibility. In the 0-2 m/s range, the violation count decreases from 826 in the noise-centric baseline to 677 in the trajectory-centric version, and further to only 8 after adaptive curvature-regularized training. In the $\geq 2$ m/s range, the count drops from 554 to 235, and then further to 8. These results further show that the proposed training improves the intrinsic geometric and kinematic feasibility of trajectories.

\textbf{Effect of drivable-area guided sampling.}
Compared with +adaptive curvature-regularized training, PDMS increases from \(86.57\) to \(88.75\), DAC improves from \(94.94\) to \(97.46\), and the drivable-area violation rate decreases from \(5.06\%\) to \(2.54\%\). Meanwhile, the curvature violation rate increases moderately from \(0.13\%\) to \(0.88\%\). This result is consistent with the intended role of the guidance module: it mainly improves the drivable-area aspect of trajectory-space feasibility by steering the generated trajectory back toward the drivable area during sampling. The same trend is reflected in Fig.~\ref{fig:drivable_violation}, where the number of drivable-area violations decreases from \(748\) in the baseline to \(658\) in the trajectory-centric model, and further to \(317\) after introducing feasibility guidance.

\textbf{Post-training with feasibility-aware GRPO.}
Table~\ref{tab:ablation_rlft} compares standard GRPO and the proposed feasibility-aware GRPO. Starting from FeaXDrive-IL, standard GRPO improves PDMS from \(88.75\) to \(90.56\), but also increases the curvature violation rate from \(0.88\%\) to \(5.79\%\). In contrast, feasibility-aware GRPO achieves a PDMS of \(90.00\) while keeping the curvature violation rate at a much lower level of \(2.40\%\). At the same time, it attains slightly higher DAC (\(98.31\) vs.\ \(98.28\)) and a slightly lower drivable-area violation rate (\(1.69\%\) vs.\ \(1.72\%\)) than standard GRPO. These results indicate that standard benchmark-oriented GRPO tends to improve score at the expense of feasibility, whereas the proposed feasibility-aware GRPO achieves a better balance between benchmark performance and trajectory-space feasibility, with only a limited reduction in PDMS.

\begin{table}[t]
\centering
\caption{Comparison of post-training fine-tuning strategies for FeaXDrive-IL. FA-GRPO denotes feasibility-aware GRPO.}
\small
\setlength{\tabcolsep}{5pt}
\renewcommand{\arraystretch}{1.08}
\begin{tabular}{l|ccc|cc}
\toprule
\textbf{Method} & \textbf{PDMS} $\uparrow$ & \textbf{EP} $\uparrow$ & \textbf{DAC} $\uparrow$ & \makecell[c]{\textbf{Drivable Area}\\ \textbf{Viol. Rate} $\downarrow$} & \makecell[c]{\textbf{Curvature}\\ \textbf{Viol. Rate} $\downarrow$} \\
\midrule
FeaXDrive-IL & 88.75 & 83.34 & 97.46 & 2.54\% & \textbf{0.88\%} \\
FeaXDrive w/GRPO & \textbf{90.56} & \textbf{85.13} & 98.28 & 1.72\% & 5.79\% \\
\makecell[l]{FeaXDrive w/FA-GRPO} & 90.00 & 84.20 & \textbf{98.31} & \textbf{1.69\%} & 2.40\% \\
\bottomrule
\end{tabular}
\label{tab:ablation_rlft}
\end{table}

\textbf{Overall analysis.}
Taken together, the ablation results reveal a clear division of roles among the proposed modules. The trajectory-centric reformulation provides a unified foundation for feasibility-aware training, inference-time geometric guidance, and post-training feasibility-aware optimization; adaptive curvature-regularized training mainly improves the intrinsic geometric and kinematic feasibility of trajectories; drivable-area guidance primarily enhances the drivable-area aspect of trajectory-space feasibility during sampling; and feasibility-aware GRPO further improves planning performance while balancing trajectory-space feasibility during post-training. Overall, the step-by-step ablation results support the design of FeaXDrive and show that the proposed method enhances trajectory-space feasibility while maintaining strong closed-loop planning performance.

\subsection{Inference Efficiency}

\begin{figure}[t]
    \centering
    \includegraphics[width=0.6\columnwidth]{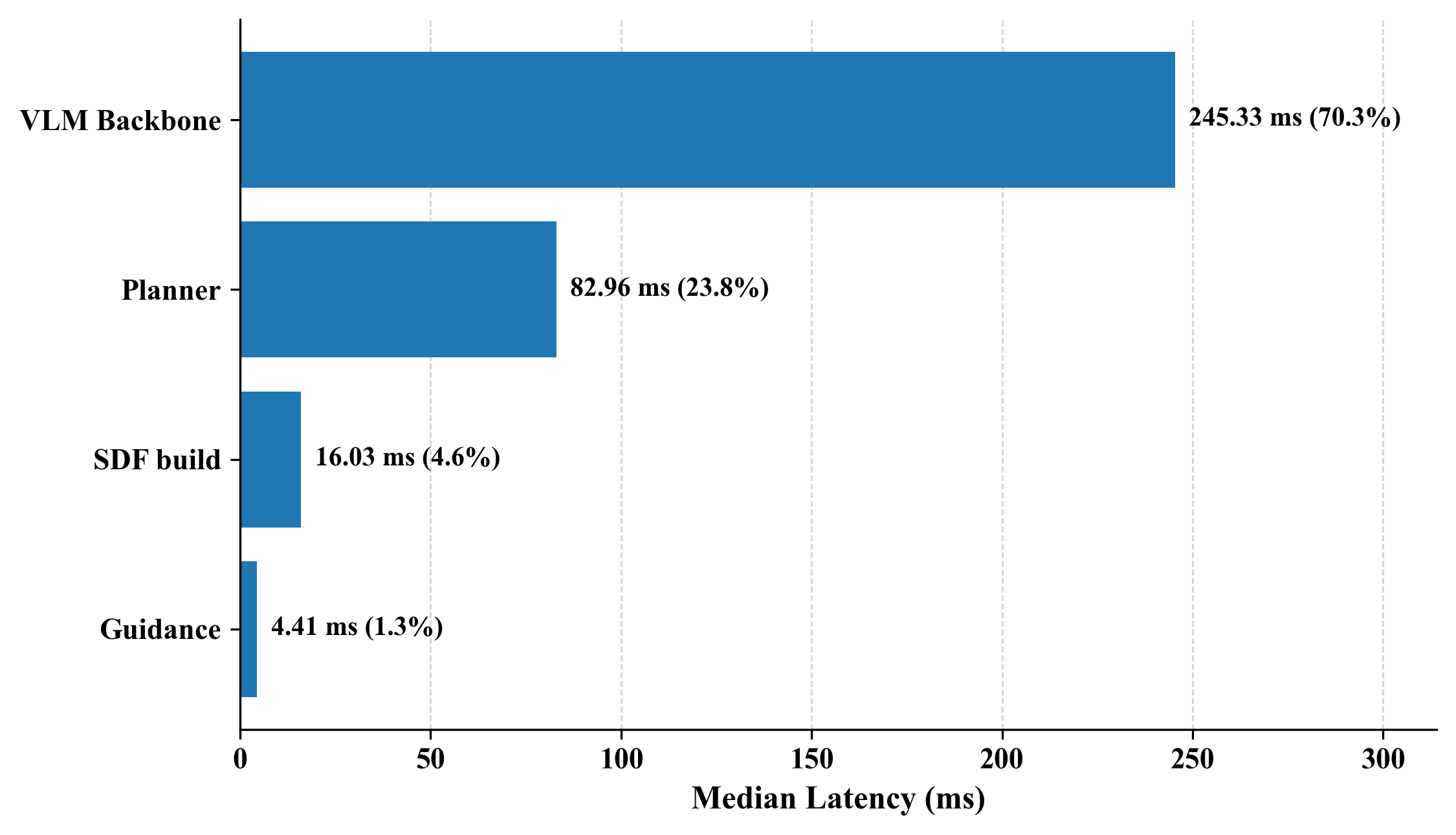}
    \caption{Latency breakdown of FeaXDrive inference.}
    \label{fig:latency_breakdown}
\end{figure}

We further analyze the inference efficiency of FeaXDrive. Fig.~\ref{fig:latency_breakdown} presents the latency breakdown of the proposed planner. Overall, the total online inference latency is \(348.73\) ms. Among all components, the VLM backbone dominates the computational cost, with a latency of \(245.33\) ms (\(70.3\%\) of the total), followed by the planner module with \(82.96\) ms (\(23.8\%\)). In contrast, the additional overhead introduced by the proposed feasibility-aware components remains small: local SDF construction costs \(16.03\) ms (\(4.6\%\)), while drivable-area guidance adds only \(4.41\) ms (\(1.3\%\)). These results indicate that the primary computational bottleneck still lies in the visual-language backbone, while the proposed feasibility-aware guidance introduces limited extra cost.

This breakdown also helps explain the practicality of the proposed method. Although FeaXDrive introduces trajectory-space feasibility modeling, its online geometric guidance remains lightweight compared with the backbone and planner. In particular, the overhead of feasibility guidance is \(4.41\) ms, which is small compared with the overall inference budget. This suggests that the proposed feasibility-aware design improves trajectory-space feasibility with modest additional online cost.

\subsection{Qualitative Results}

\begin{figure*}[t]
    \centering

    \begin{subfigure}[t]{0.48\textwidth}
        \centering
        \includegraphics[width=\linewidth]{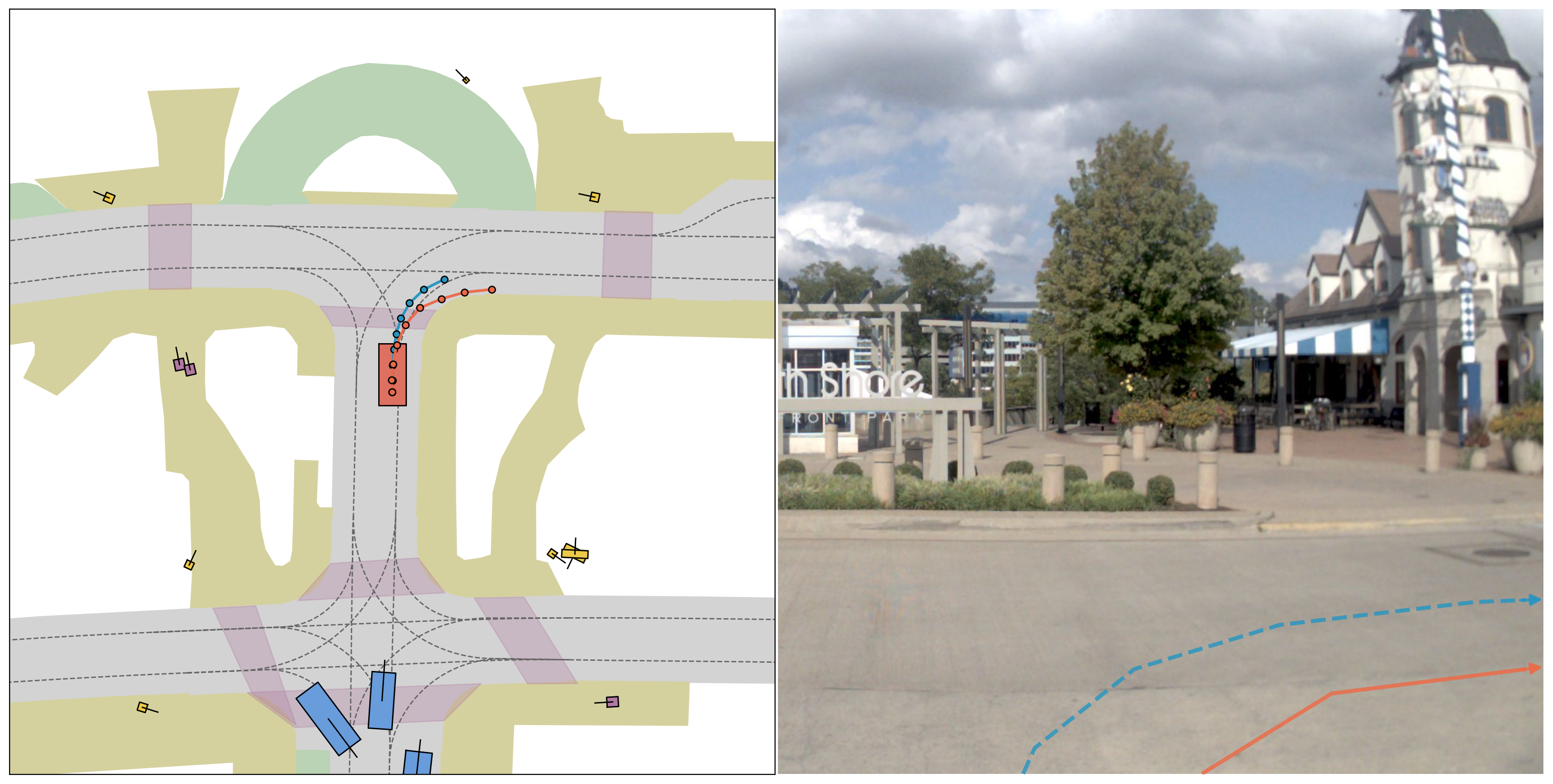}
        \caption{Baseline}
    \end{subfigure}
    \hfill
    \begin{subfigure}[t]{0.48\textwidth}
        \centering
        \includegraphics[width=\linewidth]{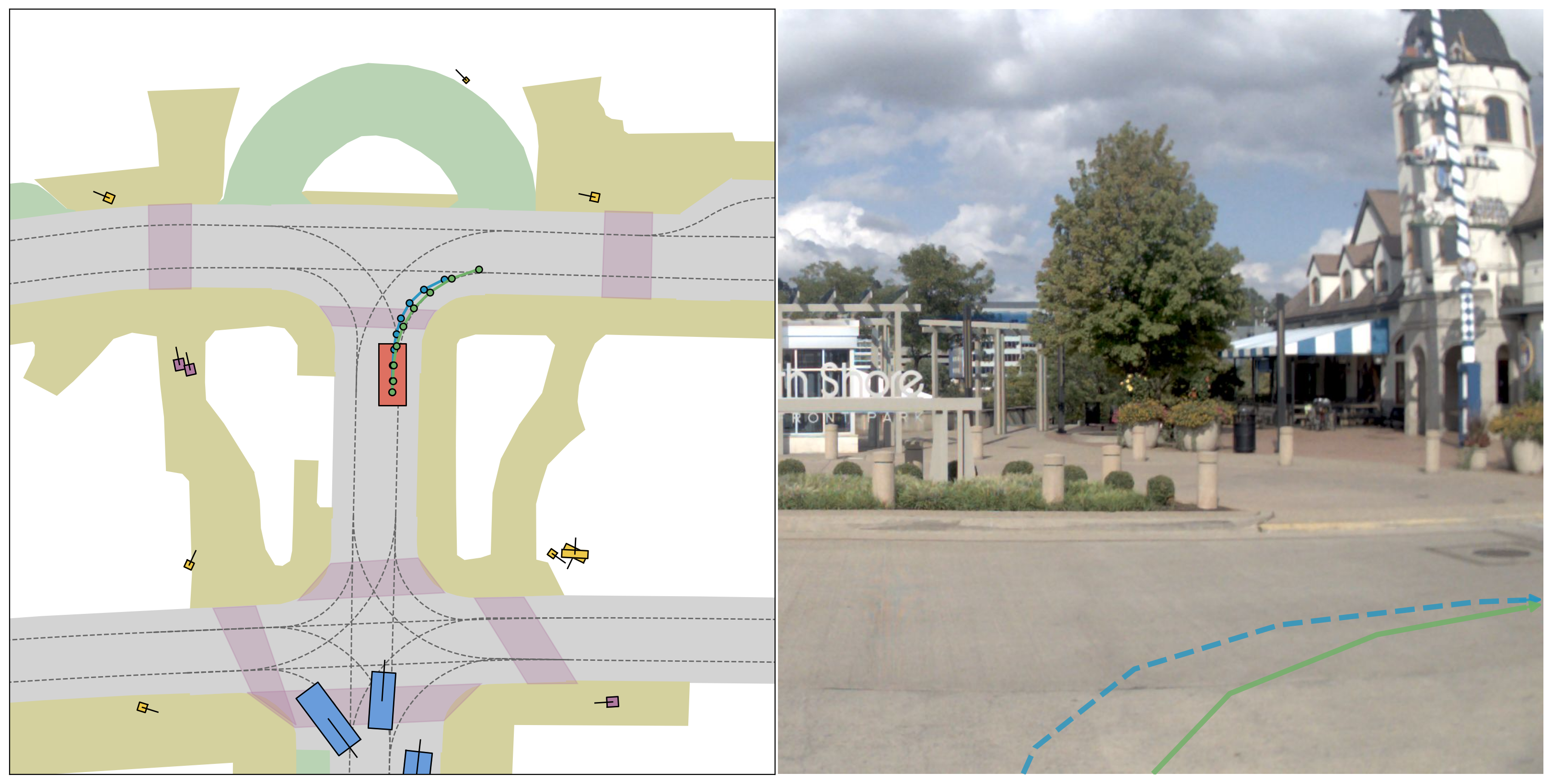}
        \caption{FeaXDrive}
    \end{subfigure}

    \vspace{0.6em}

    \begin{subfigure}[t]{0.48\textwidth}
        \centering
        \includegraphics[width=\linewidth]{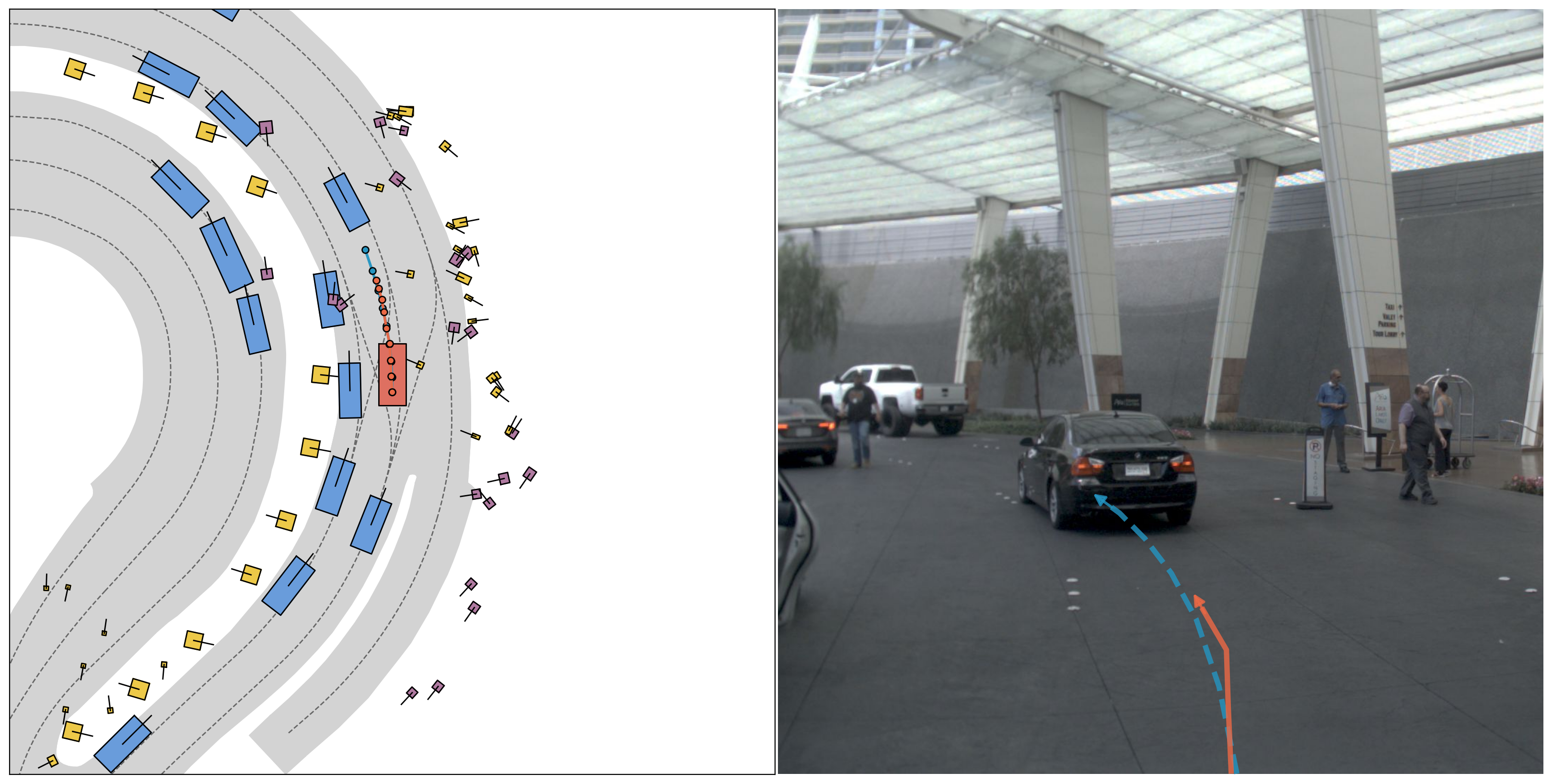}
        \caption{Baseline}
    \end{subfigure}
    \hfill
    \begin{subfigure}[t]{0.48\textwidth}
        \centering
        \includegraphics[width=\linewidth]{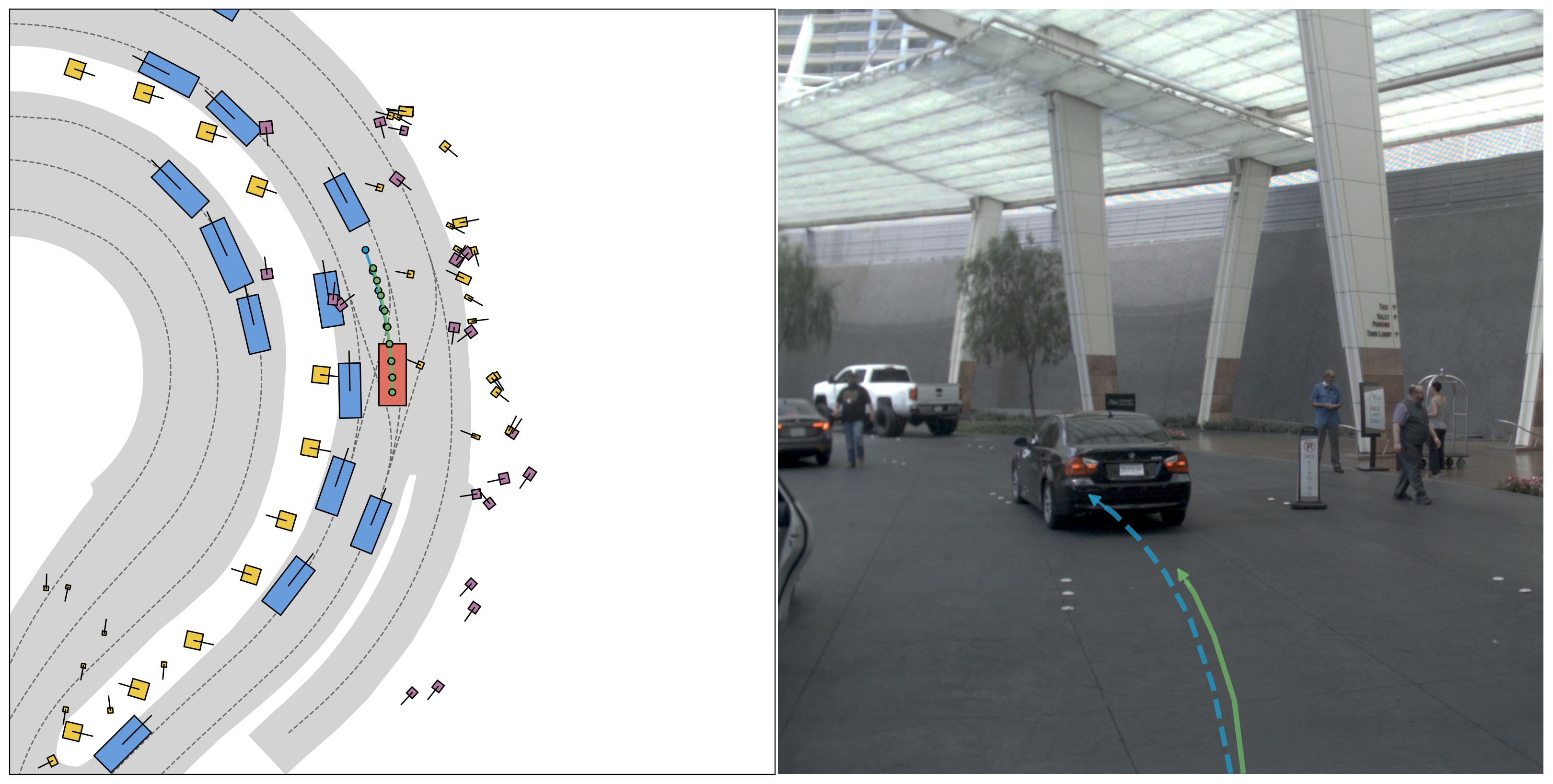}
        \caption{FeaXDrive}
    \end{subfigure}

    \vspace{0.6em}

    \begin{subfigure}[t]{0.48\textwidth}
        \centering
        \includegraphics[width=\linewidth]{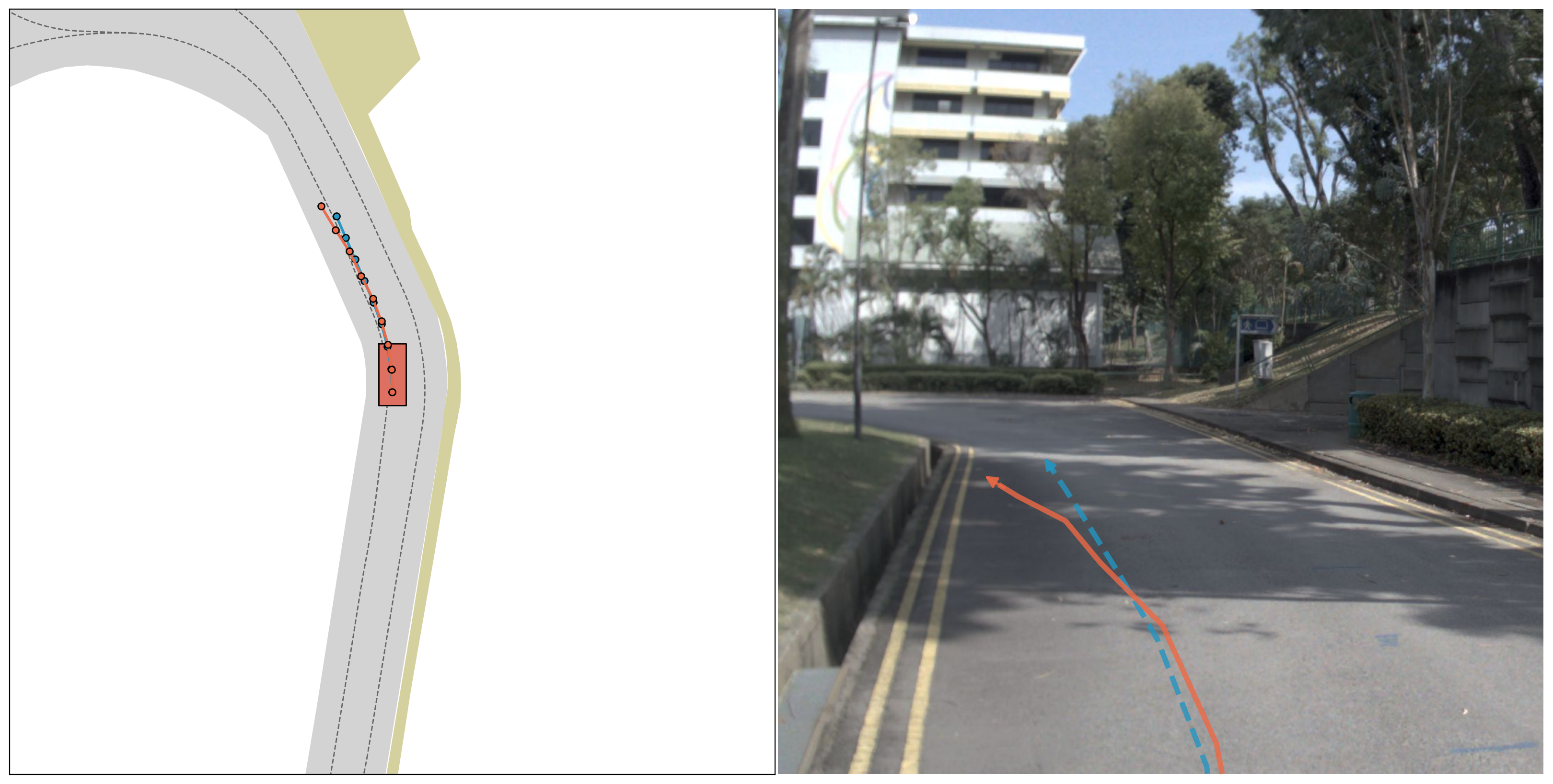}
        \caption{Baseline}
    \end{subfigure}
    \hfill
    \begin{subfigure}[t]{0.48\textwidth}
        \centering
        \includegraphics[width=\linewidth]{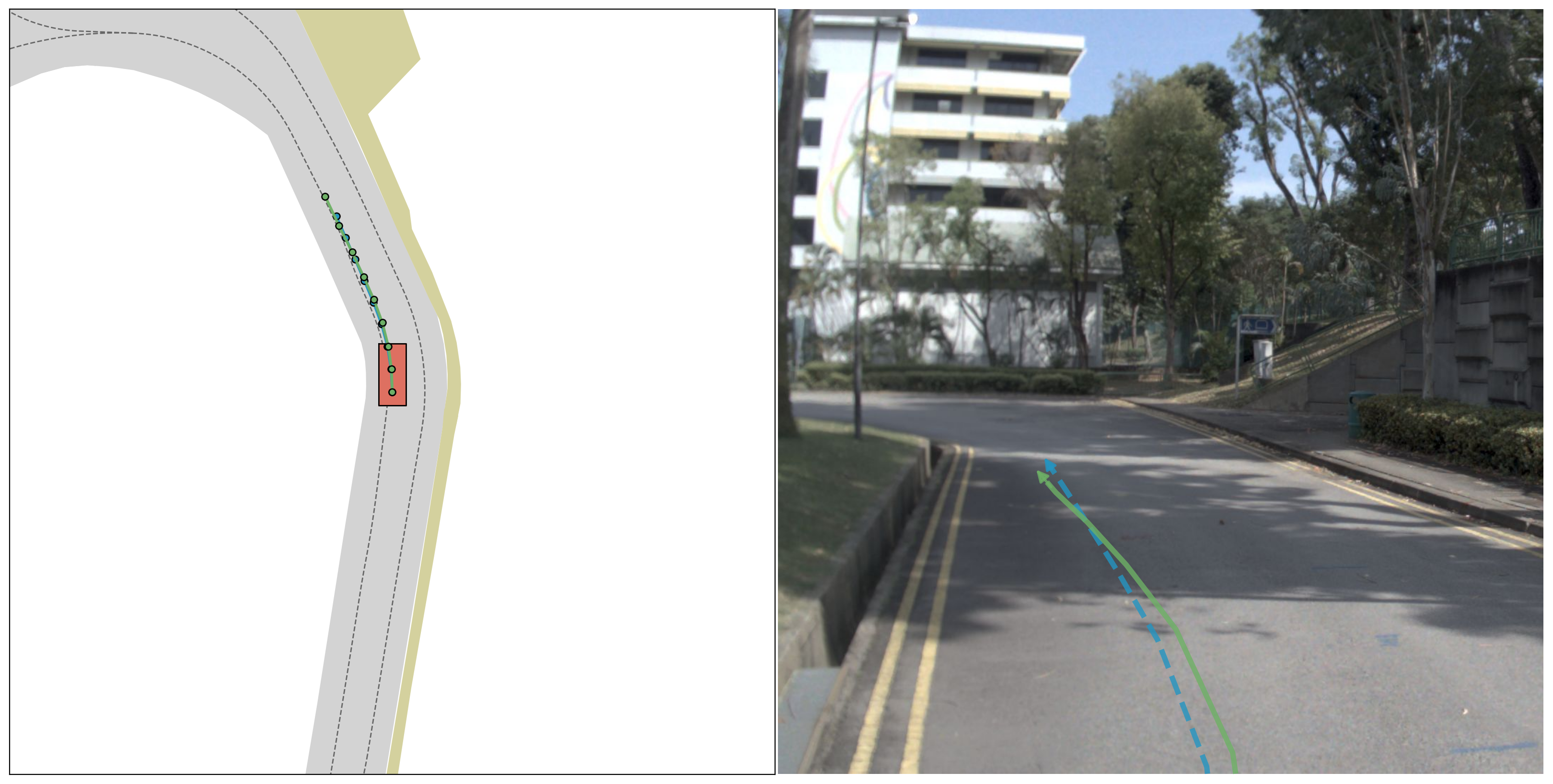}
        \caption{FeaXDrive}
    \end{subfigure}

    \caption{Qualitative comparison between the noise-centric baseline and FeaXDrive on representative planning scenes. From top to bottom: trajectory-level kinematic infeasibility, local geometric irregularities, and drivable-area inconsistency. Compared with the baseline, FeaXDrive produces trajectories that are smoother, more kinematically feasible, and better aligned with the drivable area.}
    \label{fig:qualitative_results}
\end{figure*}

We further provide qualitative comparisons between the noise-centric baseline and FeaXDrive on representative planning cases. As shown in Fig.~\ref{fig:qualitative_results}, the selected examples illustrate three typical issues closely related to trajectory-space feasibility, including trajectory-level kinematic infeasibility, local geometric irregularities, and drivable-area violation.

In the first group of examples, the baseline produces trajectories with sharp turning patterns, leading to curvature infeasibility from a trajectory-level kinematic perspective. By comparison, FeaXDrive generates trajectories with smoother turning profiles and more feasible curvature evolution, showing that the proposed trajectory-centric feasibility modeling effectively improves kinematic plausibility.

In the second group, the baseline trajectories exhibit obvious local geometric irregularities. In contrast, FeaXDrive produces smoother trajectories with more coherent local geometry. These improvements are consistent with the effect of the proposed adaptive curvature-regularized training, which suppresses local geometric spikes and improves the intrinsic geometric and kinematic feasibility of trajectories.

In the third group, the baseline trajectories deviate from the drivable area. By contrast, FeaXDrive keeps the predicted trajectory much closer to the drivable area and avoids obvious drivable-area violations. This result qualitatively verifies the effect of drivable-area guidance, which progressively corrects the clean trajectory during reverse diffusion sampling.

Overall, these qualitative examples are consistent with the quantitative results reported in the previous sections. Compared with the baseline, FeaXDrive generates trajectories with stronger trajectory-space feasibility, exhibiting better geometric regularity, improved kinematic feasibility, and better consistency with the drivable area.

\section{Discussion}

The results of this work suggest that different aspects of trajectory-space feasibility may interact in a non-trivial manner. Adaptive curvature-regularized training suppresses curvature violations, whereas drivable-area guidance further improves benchmark performance and drivable-area compliance but may moderately relax the curvature optimum achieved by training. Similarly, benchmark-oriented GRPO improves score more aggressively but sacrifices trajectory-space feasibility, while feasibility-aware GRPO preserves a better balance between planning performance and trajectory-space feasibility.

This work still has several limitations. First, the feasibility-aware GRPO stage mainly incorporates the intrinsic geometric and kinematic aspect of trajectory-space feasibility, and does not yet unify all aspects of trajectory-space feasibility within the reward. Second, drivable-area guidance currently relies on local map-derived geometric priors, although such priors could in principle also be provided by lightweight maps, online mapping, or other compact scene representations. Future work may explore more unified feasibility modeling across training, inference, and post-training, as well as lighter-weight scene priors and more efficient VLM reasoning for end-to-end autonomous driving planning.

\section{Conclusion}

This paper presented FeaXDrive, a feasibility-aware trajectory-centric diffusion planning method for end-to-end autonomous driving. Built on a trajectory-centric formulation that provides a unified foundation for feasibility-aware modeling, FeaXDrive integrates adaptive curvature-regularized training, drivable-area guidance during reverse diffusion sampling, and feasibility-aware GRPO post-training.

Experiments on NAVSIM showed that FeaXDrive achieves strong closed-loop planning performance while improving trajectory-space feasibility. The results demonstrated that the trajectory-centric formulation provides a more suitable and unified foundation for feasibility-aware modeling, while adaptive curvature-regularized training improves intrinsic geometric and kinematic feasibility, drivable-area guidance enhances consistency with the drivable area, and feasibility-aware GRPO further improves planning performance while balancing trajectory-space feasibility during post-training. Overall, this work highlights the importance of explicitly modeling trajectory-space feasibility in end-to-end diffusion planning and provides a step toward more reliable and physically grounded autonomous driving planners.

\bibliographystyle{elsarticle-harv}
\bibliography{refs}

\end{document}